\pdfoutput=1
\documentclass[journal,twoside,web]{ieeecolor}

\usepackage{etoolbox}
\makeatletter
\@ifundefined{color@begingroup}%
{\let\color@begingroup\relax
\let\color@endgroup\relax}{}%
\def\fix@ieeecolor@hbox#1{%
\hbox{\color@begingroup#1\color@endgroup}}
\patchcmd\@makecaption{\hbox}{\fix@ieeecolor@hbox}{}{\FAILED}
\patchcmd\@makecaption{\hbox}{\fix@ieeecolor@hbox}{}{\FAILED}

\usepackage{tmi}
\usepackage{cite}
\usepackage{amsmath,amssymb,amsfonts}
\usepackage{algorithmic}
\usepackage{graphicx}
\usepackage{textcomp}
\usepackage{multirow}
\usepackage{bbm}
\usepackage{booktabs}
\usepackage{color, colortbl}
\usepackage[dvipsnames]{xcolor}
\definecolor{mygray}{gray}{.9}
\usepackage{hyperref}
\usepackage{caption}

\def\BibTeX{{\rm B\kern-.05em{\sc i\kern-.025em b}\kern-.08em
    T\kern-.1667em\lower.7ex\hbox{E}\kern-.125emX}}
\begin{document}
\title{QMix: Quality-aware Learning with Mixed Noise for Robust Retinal Disease Diagnosis}
\author{Junlin Hou, Jilan Xu, Rui Feng, Hao Chen, \IEEEmembership{Senior Member, IEEE}
\thanks{This work is submitted on Aug 23rd, 2024. This work was supported by National Natural Science Foundation of China (No. 62172101), the Hong Kong Innovation and Technology Fund (Project No. PRP/034/22FX and MHP/002/22), the Project of Hetao Shenzhen-Hong Kong Science and Technology Innovation Cooperation Zone (HZQB-KCZYB-2020083), HKUST 30 for 30 Research Initiative Scheme, and the Science and Technology Commission of Shanghai Municipality (No. 23511100602). Corresponding authors: R. Feng and H. Chen. }
\thanks{J. Hou and H. Chen are with Department of Computer Science and Engineering, Hong Kong University of Science and Technology, Hong Kong, China (email: \{csejlhou, jhc\}@ust.hk). J. Xu and R. Feng are with School of Computer Science, Shanghai Key Laboratory of Intelligent Information Processing, Fudan University, Shanghai, China (email: \{jilanxu18, fengrui\}@fudan.edu.cn).
H. Chen is also with Department of Chemical and Biological Engineering and Division of Life Science, HKUST, Hong Kong, China; HKUST Shenzhen-Hong Kong Collaborative Innovation Research Institute, Futian, Shenzhen, China.}
}

\maketitle

\begin{abstract}
Due to the complex nature of medical image acquisition and annotation, medical datasets inevitably contain noise. This adversely affects the robustness and generalization of deep neural networks. Previous noise learning methods mainly considered noise arising from images being mislabeled, \emph{i.e.}, label noise, assuming all mislabeled images were of high quality. However, medical images can also suffer from severe data quality issues, \emph{i.e.}, data noise, where discriminative visual features for disease diagnosis are missing. In this paper, we propose QMix, a noise learning framework that learns a robust disease diagnosis model under mixed noise scenarios. QMix alternates between sample separation and quality-aware semi-supervised training in each epoch. The sample separation phase uses a joint uncertainty-loss criterion to effectively separate (1) correctly labeled images, (2) mislabeled high-quality images, and (3) mislabeled low-quality images. The semi-supervised training phase then learns a robust disease diagnosis model from the separated samples. Specifically, we propose a sample-reweighing loss to mitigate the effect of mislabeled low-quality images during training, and a contrastive enhancement loss to further distinguish them from correctly labeled images. QMix achieved state-of-the-art performance on six public retinal image datasets and exhibited significant improvements in robustness against mixed noise. \href{https://github.com/houjunlin/QMix}{Code will be available upon acceptance.} 

\end{abstract}

\begin{IEEEkeywords}
Label noise, Image quality, Retinal disease diagnosis, Sample separation, Semi-supervised learning
\end{IEEEkeywords}

\section{Introduction}
\begin{figure}[t]
\begin{center}
   \includegraphics[width=\linewidth]{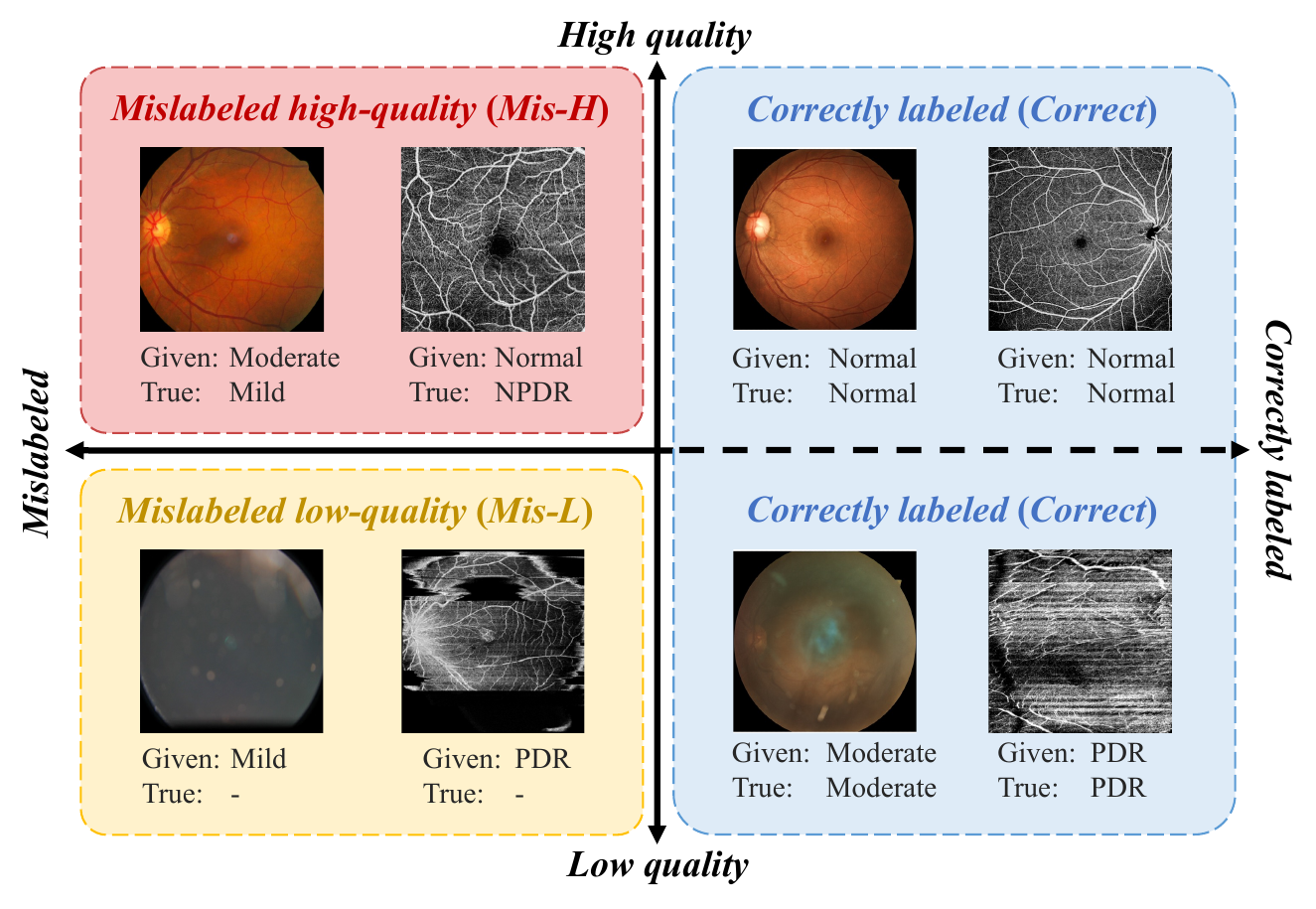}
\end{center}
\vspace{-1.5em}
   \caption{An illustration of correctly labeled data (Correct), mislabeled high-quality data (Mis-H), and mislabeled low-quality data (Mis-L). 
   }
\label{fig:intro}
\vspace{-1.0em}
\end{figure}

Deep learning has achieved significant advancements in medical image analysis. One of the main reasons is the growing availability of increasingly large datasets.
However, in practice, the complex data collection and labeling processes inherent to medical datasets often result in the presence of noise \cite{karimi2020deep}. 
For instance, doctors' labeling experience and inter-annotator variability could potentially lead to images being incorrectly labeled. 
Deep neural networks trained on such datasets are prone to over-fitting noisy labels, resulting in limited generalization ability \cite{zhang2021understanding}.
Learning from noisy labels (LNL) has been a long-standing challenge in image recognition \cite{song2022learning,algan2021image}. Most existing LNL methods aim to address this by filtering correctly labeled samples from the entire noisy dataset \cite{jiang2018mentornet,han2018coteaching,li2020dividemix}. The remaining noisy samples are either assigned smaller weights in the loss function \cite{jiang2018mentornet,guo2018curriculumnet} or treated as unlabeled data for semi-supervised learning (SSL) \cite{li2020dividemix,karim2022unicon}.

\begin{figure}[t]
\begin{center}
   \includegraphics[width=\linewidth]{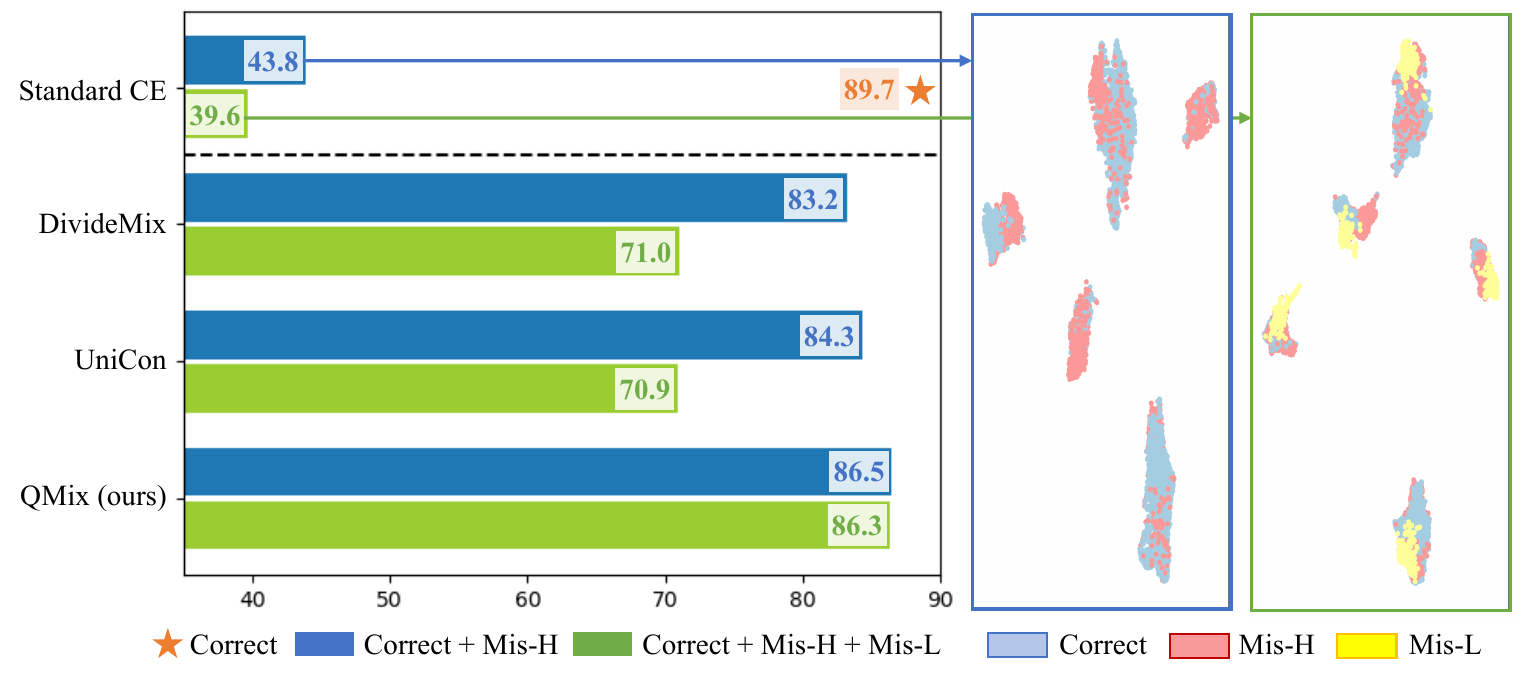}
\end{center}
\vspace{-1.0em}
   \caption{Results of DR grading on the DDR dataset. Left: performance degradation (in kappa score) of current methods under mixed noise (green bar); Right: the baseline model, trained by the standard cross-entropy loss, failed to distinguish among Correct/Mis-H/Mis-L.}
\label{fig:intro2}
\vspace{-1.0em}
\end{figure}

While existing works in the general domain have focused on addressing label noise, where the image is simply mislabeled, the sources of noise in the medical domain are much more complex. 
For example, variations in imaging conditions, differences in equipment, or improper patient cooperation can result in poor image quality.
As illustrated in Fig. \ref{fig:intro}, samples with correct disease category labels (Given = True) are valuable for network training, and we simply denote them as Correct.
Mislabeled samples (Given $\neq$ True) can be further divided into two categories: (1) mislabeled images with comparatively high image quality (Mis-H, top-left) and (2) mislabeled images with low quality (Mis-L, bottom-left) that lack sufficient information for reliable labeling.
In Fig. \ref{fig:intro2}, we empirically show that current LNL approaches \cite{li2020dividemix,karim2022unicon} managed to handle datasets containing only Correct and Mis-H (blue bars). 
However, their disease diagnosis performance dropped significantly when Mis-L were also present (green bars).
In contrast, our proposed method, QMix, exhibited robust performance under this mixed noise scenario.

In this paper, we propose a new LNL framework, termed as QMix, that learns a robust disease diagnosis model in the presence of both label noise and data noise. 
Specifically, QMix follows an alternating separate-then-learn pipeline. First, it uses a novel sample separation strategy based on a joint uncertainty-loss criterion to distinguish between Correct, Mis-H, and Mis-L.
This strategy is motivated by two observations: (1) DNNs tend to learn Correct in early training iterations by producing a relatively lowest loss, while Mis-H have a higher average loss than Mis-L; and (2) low-quality images lack discriminative visual cues, leading to higher uncertainty in network predictions.
QMix then adopts a quality-aware SSL training process to learn robust representations for disease diagnosis. The network has a dual architecture with shared parameters, \emph{i.e.}, a labeling branch that takes weakly-augmented Correct/Mis-H/Mis-L samples and generates pseudo labels, and an online branch that takes strongly-augmented counterparts and is supervised by the pseudo labels.
We optimize the network by a combination of two newly devised loss functions, including a SSL loss with sample-reweighing to mitigate the impact of Mis-L dynamically during training, and a contrastive enhancement loss to better distinguish Mis-L from Correct, benefiting the sample separation in the next round.
We evaluated our proposed QMix on six public retinal disease diagnosis benchmarks, \emph{i.e.}, DDR~\cite{li2019diagnostic}, DRTiD~\cite{hou2022cross}, DeepDRiD~\cite{deepdrid}, EyeQ~\cite{fu2019evaluation}, DRAC~\cite{qian2024drac}, and ODIR~\cite{odir2019peking}. Experimental results show that our method achieved state-of-the-art performance on all the datasets. Especially when the noise ratio is extremely high, \emph{e.g.}, 80\% Mis-H and 15\% Mis-L data, the baseline method suffered noise while QMix exhibited the robustness (15.9\% vs. 75.9\% Quadratic Weighted Kappa).  
The main contributions are listed below: 
\begin{itemize}
    \item We propose QMix, a novel LNL framework that learns a robust disease diagnosis model in the presence of mixed noise, including both label noise and data noise. 
    \item We develop a novel sample separation strategy that can distinguish between Correct, Mis-H, and Mis-L by modeling the joint distribution of uncertainty and loss. 
    \item We introduce two loss functions for robust network training, \emph{i.e.}, a sample-reweighing loss to improve quality-aware learning and a contrastive enhancement loss to encourage better discrimination between Correct and Mis-L.
    \item Extensive experiments on both synthetic and real-world noise from six retinal disease datasets show that QMix significantly outperformed other LNL methods and revealed strong robustness even at a high noise ratio.
\end{itemize}

\section{Related Work}

\subsection{Learning with Noisy Labels}

Existing methods of learning with noisy labels can be broadly categorized into two groups. The first category of works ~\cite{patrini2017making,hendrycks2018using,cheng2022instance,cheng2022class} aims to estimate a noise transition matrix, which represents the probabilities of transitioning from true labels to noisy labels, for label correction or loss calibration.
However, it is challenging to calculate the transition matrix when the number of classes grows in real-world applications.

The second category, namely sample selection, is to filter correctly labeled instances from a noisy dataset and only use them to train DNNs. Several works \cite{jiang2018mentornet,guo2018curriculumnet} discarded or reweighed noisy samples in the objective function. 
However, they failed to leverage the visual information contained in noisy samples.
Another line of works rejected original labels by relabeling all samples using network predictions \cite{tanaka2018joint} or learned label distributions \cite{yi2019probabilistic}. Malach \emph{et al}. \cite{malach2017decoupling} and Han \emph{et al}. \cite{han2018coteaching} 
adopted a co-training strategy of two networks' collaboration to alleviate error accumulation caused by incorrect selections. More recently, a group of approaches \cite{li2020dividemix,nishi2021augmentation} used SSL algorithms, where noisy samples were treated as unlabeled data. This allowed rejecting their labels while still exploiting the image content.
Combinations of SSL and contrastive learning \cite{ortego2021multi,karim2022unicon,liu2024plremix} have also been explored to combat the memorization of noisy labels.
Inspired by these works, we particularly devise a novel sample separation strategy based on an uncertainty-loss criterion to effectively distinguish between Correct, Mis-H, and Mis-L.

In addition, prior works explored using model uncertainty to help identify hard clean samples \cite{sun2020crssc} or out-of-distribution label noise \cite{shin2020strategies}. 
Similarly, model uncertainty can naturally function as an appropriate metric to differentiate low-quality samples.
However, the purpose and approach differ between our method and other approaches. 
In QMix, we model the joint uncertainty-loss distribution to separate Correct, Mis-H, and Mis-L. In contrast, other approaches like FOCI \cite{shin2020strategies} used uncertainty to assign weights to clean samples, and sample selection of clean and noisy data still relied on loss only.

\begin{figure*}[t]
\begin{center}
   \includegraphics[width=\linewidth]{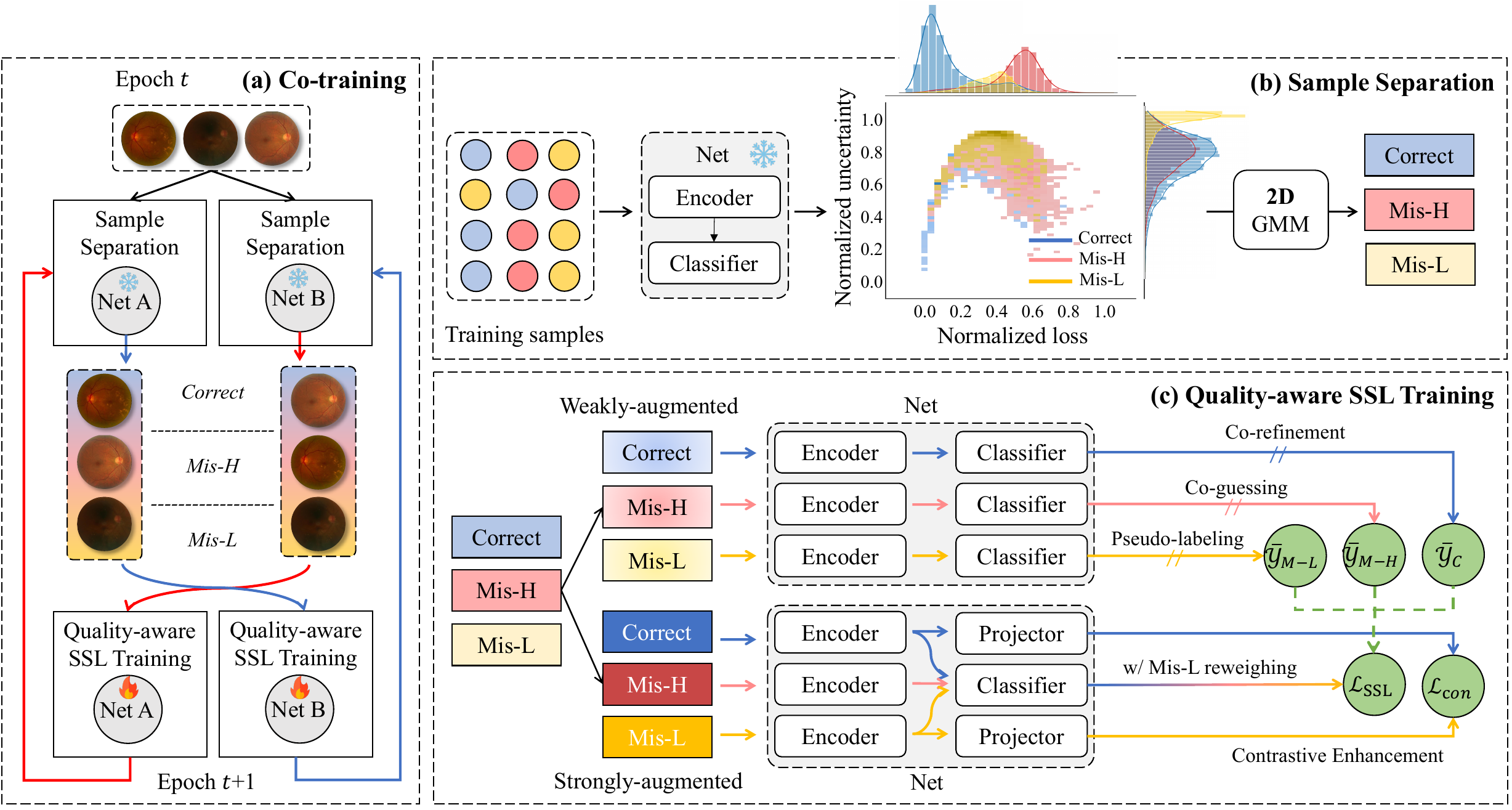}
\end{center}
\vspace{-1.0em}
   \caption{
   An overall framework of QMix for quality-aware learning with mixed noise. 
   (a) A network co-training scheme that alternates between sample separation and quality-aware semi-supervised learning (SSL) training in each epoch.
   (b) Sample separation of Correct/Mis-H/Mis-L using the joint distribution of uncertainty and loss. (c) Quality-aware SSL training with a SSL loss ($\mathcal{L}_{SSL}$) and a contrastive enhancement loss ($\mathcal{L}_{con}$).}
\label{fig:noisy_framework}
\vspace{-1.0em}
\end{figure*}

\subsection{Medical Image Analysis with Noisy Labels}

There have been several works attempting to learn robust medical image representations with noisy labels \cite{dgani2018training,le2019pancreatic,xue2019robust,li2023learning,xing2023gradient,khanal2023improving,ju2022improving,zhou2023combating}. For instance, Dgani \emph{et al}. \cite{dgani2018training} used a noise adaptation layer \cite{goldberger2016training} to estimate the transition matrix for mammography classification.
Based on sample reweighing \cite{ren2018learning}, Le \emph{et al}. \cite{le2019pancreatic} trained the model on a large noisy corpus of patches, incorporating weights from a small clean set for pancreatic cancer detection.
Similarly, Xue \emph{et al}. \cite{xue2019robust} adopted a reweighing strategy to remove samples with high loss values for skin lesion diagnosis. 
Targeting classification in imbalanced noisy datasets, Li \emph{et al}. \cite{li2023learning} proposed a multi-stage noise removal framework that simultaneously addresses the negative impacts of label noise, imbalanced distributions, and class hardness.
Additionally,
Xing \emph{et al}. \cite{xing2023gradient} enhanced the two-stage framework for sample selection and SSL training by incorporating gradient and feature conformity-based selection, along with a sample reliability-based mixup method.
Furthermore, Khanal \emph{et al}. \cite{khanal2023improving} investigated contrastive and pretext task-based self-supervised pretraining to initialize model weights for handling self-induced noisy labels.
For retinal disease classification, Ju \emph{et al}. \cite{ju2022improving} used dual-uncertainty estimation to tackle disagreement and single-target label noise, while
Zhou \emph{et al}. \cite{zhou2023combating} employed consistency regularization and disentangled distribution learning to combat noisy labels.
However, most works directly adapted existing LNL methods from general domain to medical domain, without developing specialized methods to meet the characteristics of medical noisy labels. 
In this work, we dive deeper into the issue of mislabeled images and consider the image quality problem that frequently occurs in medical domain. We propose effective sample separation and respective LNL training strategies to address these challenges.

\section{Methodology}
In this section, we introduce our proposed LNL framework, termed as QMix, which is comprised of sample separation and quality-aware semi-supervised learning (SSL) training. 
As illustrated in Fig. \ref{fig:noisy_framework}, these two processes \emph{execute alternatively} in each training epoch.    
Sample separation (Sec. \ref{sec:ss}) aims to distinguish the input data into Correct, Mis-H, and Mis-L via a newly designed uncertainty-loss criterion.
Subsequently, the quality-aware SSL training process (Sec. \ref{sec:ssltraining}) trains the network to learn robust feature representations using a novel SSL loss with sample-reweighing and a contrastive enhancement loss. 
To avoid error accumulation in self-training, QMix employs a co-training scheme, where the training data separated by each network is used to train the other network.

\subsection{Sample Separation} \label{sec:ss}

Given the training data $\mathbbm{D}=\{(x_i,y_i)\}_{i=1}^N$ where $x_i$ is an image and $y_i\in \{0,1\}^C$ is the one-hot label over $C$ disease classes. 
Sample separation aims to identify the possible noisy samples from the correctly labeled data by assigning a noise indicator $n_i\in \{0,1,2\}$, where 0, 1, and 2 refer to correctly labeled images (Correct), mislabeled high-quality images (Mis-H), and mislabeled low-quality images (Mis-L), respectively. Notably, the separation is conducted in an unsupervised manner, \emph{i.e.}, the ground-truth label of samples being Correct/Mis-H/Mis-L is not available in practice. 
We start by reviewing the basic memorization effect, which has been widely used to guide the separation between Correct and Mis-H. Then, we introduce our proposed solution to address the more challenging sample separation where both Mis-H and Mis-L are involved.

\subsubsection{Memorization effect on mixed label noise}
\begin{figure}[t]
\begin{center}
   \includegraphics[width=\linewidth]{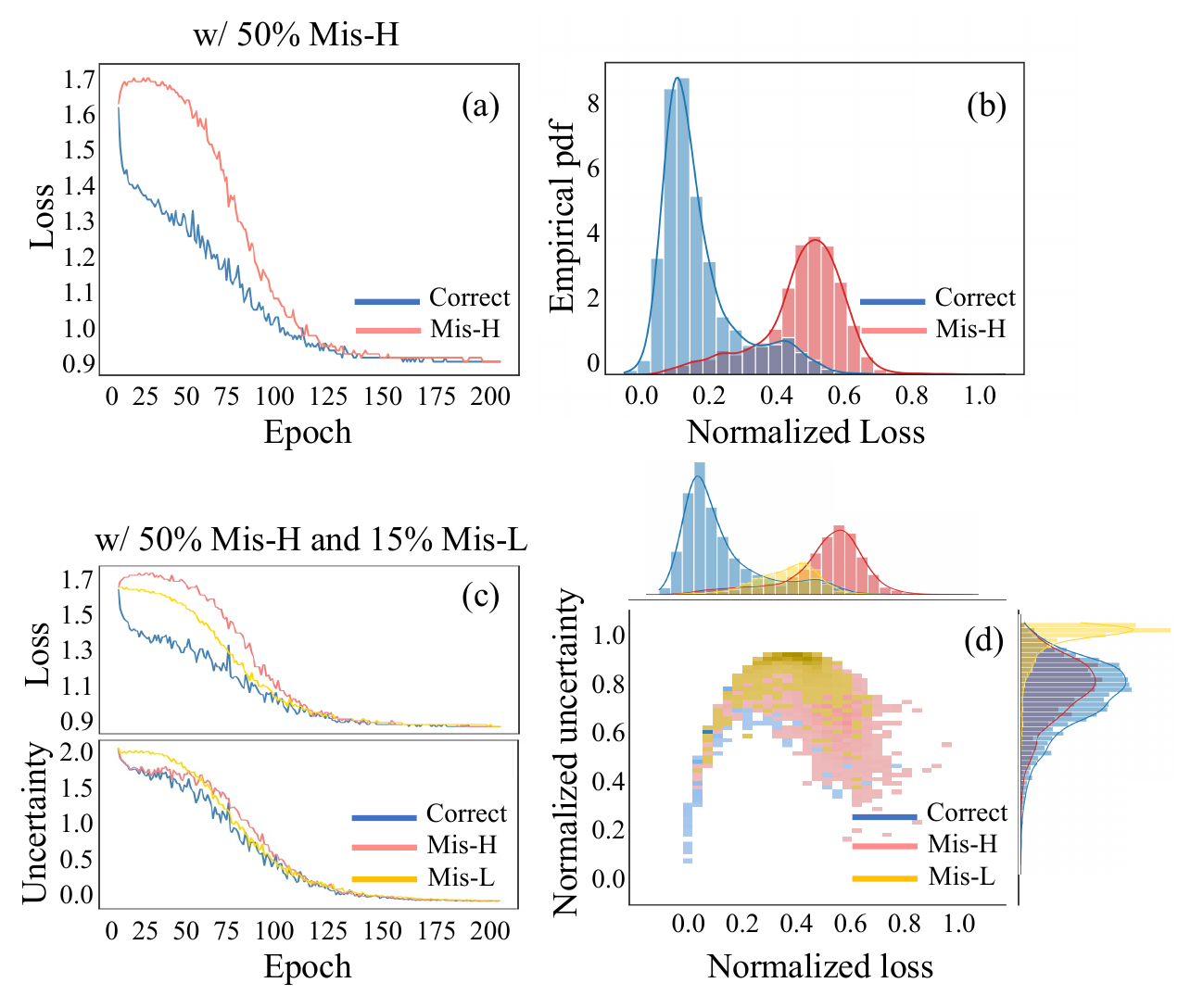}
\end{center}
\vspace{-1.0em}
   \caption{Memorization effect on (a) label noise only and (c) mixed noise; (b) previous small-loss separation; (d) our joint uncertainty-loss criterion.}
\label{fig:memorization}
\vspace{-1.0em}
\end{figure}

Memorization effect \cite{arpit2017closer} refers to the behavior exhibited by DNNs trained on noisy samples. 
As shown in Fig. \ref{fig:memorization}(a), the network first learns simple patterns from Correct in the early training iterations. As the training process continues, the network gradually memorizes Mis-H by over-fitting wrong labels. 
Prior works \cite{jiang2018mentornet,arazo2019unsupervised,han2018coteaching,yu2019does,li2020dividemix} leveraged this memorization effect to treat samples with smaller losses as Correct and selected them from Mis-H, as shown in Fig. \ref{fig:memorization}(b). While effective, these approaches only distinguish between Correct and Mis-H, \emph{i.e.}, $n_i\in\{0, 1\}$, without considering Mis-L ($n_i=2$) in the dataset.
We further investigate the memorization effect in the presence of mixed noise, where both Mis-H and Mis-L are included. 
As can be observed in Fig. \ref{fig:memorization}(c), Mis-H maintains a higher average loss in the early stage of training. This is because the model initially learns from Correct, enabling it to generate correct predictions for Mis-H. 
However, incorrect labels lead to a higher loss for Mis-H.
In contrast, Mis-L lacks distinctive features, causing model's predictions to be evenly distributed. Consequently, the average loss for Mis-L is lower compared to Mis-H.
Nevertheless, the small-loss criterion initially designed to separate Correct from Mis-H fails to handle mixed noise, as the loss distribution of Mis-L may have overlap with Correct and Mis-H as shown in Fig. \ref{fig:memorization}(d) top.

\subsubsection{Uncertainty-loss criterion for sample separation}

Generally, both Correct and Mis-H provide clear visual patterns for the network to make accurate predictions in the early stage of training. In contrast, Mis-L lacks discriminative visual information, leading to higher uncertainty in network predictions, as shown in Fig. \ref{fig:memorization} (c) bottom. This motivates us to conduct sample separation based on the joint distribution of per-sample uncertainty and loss, as depicted in Fig. \ref{fig:memorization}(d). 
Samples exhibiting small loss and low uncertainty are classified as Correct. Meanwhile, Mis-L (with moderate loss and high uncertainty) and Mis-H ( with high loss and low uncertainty) can also be separated accordingly.

Formally, given the training samples $\mathbbm{D}=\{(x_i,y_i)\}_{i=1}^N$, we train a disease classification model $F(\phi)$ parameterized by $\phi$. 
We adopt the standard cross-entropy loss $l(\phi)$ to reflect how well the model fits the training samples and the entropy $e(\phi)$ to measure the uncertainty of the model's predictions:
\begin{equation}
\begin{aligned}
    (l(\phi),e(\phi))&=\{(l_i,e_i)\}_{i=1}^N\\
    &=\{(-y_i\log \mathrm{p}(x_i;\phi),-\mathrm{p}(x_i;\phi) \log \mathrm{p}(x_i;\phi))\}_{i=1}^N,    
\end{aligned}
\end{equation}
where $\mathrm{p}(x_i;\phi)$ denotes the model's output softmax probability.

To separate Correct, Mis-H, and Mis-L, we fit a three-component Gaussian Mixture Model (GMM) \cite{richardson1997bayesian} to the joint distribution of normalized $(l(\phi),e(\phi))$. 
In specific, we define the model for each sample $s_i=(l_i,e_i)$ as a probability mixture model of three 2D Gaussian distributions $\mathcal{N}(\mu^1,\mu^2,\sigma^1, \sigma^2, \rho)$, where $\mu^1$ and $\sigma^1$ denote the mean and variance of the loss variable $l(\phi)$, $\mu^2$ and $\sigma^2$ denote the mean and variance of the uncertainty variable $e(\phi)$, $\rho$ indicates the correlation between loss and uncertainty:
\begin{equation}
    \mathrm{p}(s_i|\theta)=\sum\nolimits_{k=0}^2 \lambda_k \mathcal{N}(s_i|\theta_k).
\end{equation}
Here, $\theta_k=(\mu^1_k,\mu^2_k,\sigma^1_k, \sigma^2_k, \rho_k)$ denotes the parameter of the $k$-th 2D gaussian distribution; $\theta=\{\theta_k,\lambda_k\}_{k=0,1,2}$.
We adopt the Expectation-Maximization algorithm \cite{dempster1977maximum} to obtain the posterior probability $w_i^k$ of sample $s_i$ belonging to the $k$-th Gaussian component $z_k$:
\begin{equation}
    w^k_i = \mathrm{p}(z_k|s_i)=\frac{\lambda_k \mathcal{N}(s_i|\theta_k)}{\sum_{j=0}^2 \lambda_j \mathcal{N}(s_i|\theta_j)}
\label{eq:w}
\end{equation}
The predicted noise indicator $n_i$ is computed by indexing the maximum probability over three Gaussian components:
\begin{equation}
n_i = \text{argmax}_{k\in\{0,1,2\}} w^k_i.  
\end{equation}

\noindent\textbf{Network Co-training.}
The accuracy of sample separation has a significant impact on the subsequent SSL training. For example, if Mis-H is incorrectly classified as Correct, it would keep having lower loss due to the model over-fitting to its label.
To avoid such error accumulation, we adopt a co-training strategy \cite{malach2017decoupling}, as illustrated in Fig. \ref{fig:noisy_framework}(a).
Specifically, two networks, $F(\phi_A)$ and $F(\phi_B)$, are trained simultaneously in each epoch. 
The samples $\{(x_i,y_i,n_i^A)\}_{i=1}^N$ with their noise indicators $n_i^A$ produced by $F(\phi_A)$ are used to guide the SSL training for the other network $F(\phi_B)$, and vice versa. 
This co-training scheme encourages the two networks to diverge from each other and become more robust to noise, due to different initialization and data divisions \cite{li2020dividemix}.
Notably, the co-training scheme is only adopted during the training phase. In the test phase, either $F(\phi_A)$, $F(\phi_B)$, or an ensemble of $F(\phi_A)$ and $F(\phi_B)$ can be used for robust disease diagnosis. By default, we adopt the ensemble of both networks, and a comparative analysis can be found in Sec. \ref{subsec:ablation}.

\subsection{Quality-aware Semi-Supervised Training} \label{sec:ssltraining}
In this phase, we aim to train the classification network to learn robust feature representations from these separated samples through a novel semi-supervised learning (SSL) pipeline. 
As shown in Fig. \ref{fig:noisy_framework}(c), we introduce a dual architecture with shared model parameters. The labeling branch takes the weakly-augmented Correct, Mis-H, and Mis-L as input and generates pseudo disease diagnosis labels to supervise the online branch. 
The online branch inputs the strongly-augmented samples and is trained by optimizing the proposed combined loss. 
Note that, only the online branch is trained, while the labeling branch always maintains the same model parameters as the online branch. 
Given the training samples $\mathbbm{D}=\{(x_i,y_i,w_i,n_i)\}_{i=1}^N$, where $w_i=\{w_i^k\}_{k=0}^2$ refers to the probability of the sample being Correct/Mis-H/Mis-L produced by GMM and $n_i$ denotes its predicted noise indicator. 
Based on the output of GMM, Correct samples $\mathbbm{D}_{correct}=\{(x_i,y_i,w_i, n_i)| n_i=0\}_{i=1}^{N_c}$ are regarded as labeled samples, Mis-H samples $\mathbbm{D}_{mis-h}=\{(x_i,w_i, n_i)| n_i=1\}_{i=1}^{N_h}$ are treated as unlabeled samples by discarding their original class labels,
while Mis-L samples $\mathbbm{D}_{mis-l}=\{(x_i,y_i,w_i, n_i)| n_i=2\}_{i=1}^{N_l}$ are \textit{progressively discarded} during the training process.

\subsubsection{Labeling branch}

For each labeled sample $x\in\mathbbm{D}_{correct}$, we refine its classification label using a co-refinement strategy.
Specifically, we linearly combine (1) the original label $y$ and (2) the average predicted class probability of $M$ weakly-augmented versions of $x$:
\begin{equation}
    \bar{y}=wy + (1-w)\frac{1}{M}\sum\nolimits_{m=1}^M \mathrm{p}(g_{weak}(x);\phi).
    \label{eq:co-refinement}
\end{equation}
Here, $w$ represents the weight of retaining the original label, with choices of specific values available in Sec. \ref{subsec:ablation} Table \ref{table:ablation_weight}; $g_{weak}$ refers to a weak-augmentation function. 

For an unlabeled sample $x\in \mathbbm{D}_{mis-h}$, we perform co-guessing to predict the pseudo label by integrating the predictions of $2M$ weakly-augmented samples from both networks:
\begin{equation}
    \bar{y}=\frac{1}{2M}\sum\nolimits_{m=1}^{2M} \mathrm{p}(g_{weak}(x);\phi).     
\end{equation}
In our experiments, we set $M=2$.

For each sample $x\in\mathbbm{D}_{mis-l}$, we gradually remove it during the training process by decreasing the loss weight to 0. The pseudo label is assigned in the same form as Eq. (\ref{eq:co-refinement}).

\subsubsection{Online branch training}
Next, we conduct quality-aware SSL training based on the strongly-augmented samples. 
Given the samples with refined labels $\{x_i,\bar{y}_i\}_{i=1}^{N}$, we adopt different loss functions for Correct, Mis-H, and Mis-L. 
Specifically, we use the standard cross-entropy (CE) loss $\mathcal{L}_{correct}$ on Correct and the mean square error (MSE) $\mathcal{L}_{mis-h}$ on Mis-H, as MSE is less sensitive to incorrect pseudo labels \cite{berthelot2019mixmatch}. Here, $g_{strong}(x)$ indicates a strong-augmentation function:

\begin{gather}
    \mathcal{L}_{correct}=-\frac{1}{|\mathbbm{D}_{correct}|}\sum_{x\in \mathbbm{D}_{correct}}\bar{y} \log (\mathrm{p}(g_{strong}(x);\phi)),\\
    \mathcal{L}_{mis-h} = \frac{1}{|\mathbbm{D}_{mis-h}|}\sum_{x\in \mathbbm{D}_{mis-h}} \left\|\bar{y}-\mathrm{p}(g_{strong}(x);\phi)\right\|^2_2.    
\end{gather}
For Mis-L, we particularly propose a sample-reweighing loss  to mitigate their potential negative impact on network training:
\begin{equation}
    \mathcal{L}_{mis-l}=-\frac{\Omega(t)}{|\mathbbm{D}_{mis-l}|}\sum_{x\in \mathbbm{D}_{mis-l}} \bar{y} \log (\mathrm{p}(g_{strong}(x);\phi)),
\end{equation}
\noindent where $\Omega(t)$ denotes the weight over Mis-L at epoch $t$.
Since Mis-L contains less distinguishable visual information, an intuitive idea is to directly discard it by setting the weight $\Omega(t)$ to a constant 0. This ensures that Mis-L does not contribute to the loss propagation during training.
However, Correct might be incorrectly classified as Mis-L due to the insufficient model capability in the early stage of network training. 
To mitigate the potential information loss, we further improve our sample-reweighing loss by introducing a linear weight decay strategy: 
\begin{equation}
    \Omega(t)=\frac{\Omega(0)}{T_w-T}\cdot(t-T),
\end{equation}
where $T$ and $T_w$ denote the total number of training epochs and warm-up epochs, respectively. 
This strategy involves initially assigning a small weight $\Omega(0)$ to Mis-L. 
As the training progresses, the network's accuracy in classifying Correct, Mis-H, and Mis-L improves steadily. 
Concurrently, the weight $\Omega(t)$ gradually decays to 0, minimizing the adverse influence of Mis-L on network training.
Our method offers a significant advantage over traditional image quality assessments conducted before training, as it can identify Mis-L during the training process in an unsupervised manner. This eliminates the requirement for additional supervision from image quality labels. The identified Mis-L will be removed, ensuring they do not contribute to model training.

\subsubsection{Contrastive enhancement on low-quality samples}
Generally, the ground-truth label of Mis-L should not correspond to any of the $C$ disease classification categories, but rather belong to a distinct category referred to as ``ungradable''. 
Therefore, we incorporate a contrastive enhancement loss into the SSL training process, aiming to better distinguish Mis-L from Correct.
Specifically, for a strongly-augmented sample $x_i$, it is encoded and mapped to a low-dimensional vector $z_i=P(F(g_{strong}(x_i)|\phi))\in \mathbb{R}^{d_p}$, where $P(\cdot)$ is the projection network. 
We define any two samples $i$ and $j$ from the same category as positive, \emph{i.e.}, $\tilde{y}_i =\tilde{y}_j$, otherwise they are negative. Note that the total number of categories is $C+1$ ($C$ disease classes with an additional ungradable class). 
Specifically, we adopt the original label $\tilde{y}\in\{0,…,C-1\}$ for Correct and assign $\tilde{y}=C$ for Mis-L.
The contrastive enhancement loss can thus be defined as:
\begin{equation}
\begin{aligned}
        \mathcal{L}_{con}=&\frac{1}{N}\sum_{i=1}^{N} \frac{-1}{N_{\tilde{y}_i}-1}\sum_{j=1}^{N}\mathbbm{1}_{i\ne j}\cdot\mathbbm{1}_{\tilde{y}_i=\tilde{y}_j}\\
        &\cdot\log\frac{\exp(z_i^T\cdot z_j/\tau)}{\sum_{k=1}^{2N}\mathbbm{1}_{i\ne k}\cdot\exp(z_i^T\cdot z_k/\tau)},
\end{aligned}
\label{eq:infonce}
\end{equation}
where $\mathbbm{1}\in\{0,1\}$ is the indicator function; $\tau>0$ denotes the temperature; $N_{\tilde{y}_i}$ is the number of samples sharing the same label with $\tilde{y}_i$ in each mini-batch. Eq. (\ref{eq:infonce}) enforces Mis-L to be grouped together and kept separated from Correct, while the distance between different disease classes remains. The discrimination between Mis-L and Correct encourages the sample separation process in the next round.

\subsubsection{Loss function}
To ensure stable training, we implement a regularization term \cite{tanaka2018joint} that utilizes a prior distribution to regularize the model's output across all samples in the mini-batch. This prior distribution is based on the categorical distribution of the dataset, defined as $\pi_c=N_c/N$, where $N_c$ denotes the number of samples in class $c$ and $N$ is the total number of samples.
\begin{equation}
    \mathcal{L}_{reg}=\sum_c \pi_c \log \left(\pi_c / \frac{1}{|\mathbb{D}|}\sum_{x\in\mathbbm{D}}\mathrm{p}^c(x;\phi) \right).
\end{equation}

The total loss is composed of the CE loss on Correct, the sample-reweighing loss on Mis-L, the MSE loss on Mis-H, the contrastive enhancement loss, and the regularization term with balancing weights $\lambda_m, \lambda_c$:
\begin{equation}
\mathcal{L}=\underbrace{\mathcal{L}_{correct}+\mathcal{L}_{mis-l}+ \lambda_m \mathcal{L}_{mis-h}}_{\mathcal{L}_{SSL}}+\lambda_c \mathcal{L}_{con} +\mathcal{L}_{reg}.
\end{equation}

\section{Experiments}

In this section, we evaluate our QMix in different label noise scenarios, including (1) a synthetic noisy label dataset (\emph{i.e.}, DDR) and (2) real-world noisy label datasets (\emph{i.e.}, DRTiD, DeepDRiD, EyeQ, DRAC, and ODIR). We introduce these datasets (Sec.~\ref{subsec:dataset}) and experimental setups (Sec.~\ref{subsec:implement}), followed by the comparison results (Sec.~\ref{subsec:synthetic}, \ref{subsec:realworld}, and \ref{subsec:sampleseparation}) and ablation studies (Sec.~\ref{subsec:ablation}).

\begin{table}[h]
\begin{center}
\caption{Details of noisy datasets, including numbers of class, numbers of training data, numbers (ratio) of Mis-H and Mis-L training data, and numbers of testing data, syn. denotes synthetics.} 
\label{table:dataset}
\resizebox{1\columnwidth}{!}{
\begin{tabular}{lc|ccc|c}
\toprule
Dataset & \#Class & \#Train & \#Mis-H & \#Mis-L & \#Test\\
\midrule
\multicolumn{4}{l}{\textit{Synthetic noisy dataset}}\\
DDR \cite{li2019diagnostic} & 5 & 6,260 & syn. & 1,151+syn. & 2,503\\
\midrule
\multicolumn{4}{l}{\textit{Real-world noisy dataset}}\\
DRTiD \cite{hou2022cross} & 5 & 2,698 & 850 (31\%)& 698 (26\%) & 1,100\\
DeepDRiD \cite{deepdrid} & 5 &  1,200 & $\checkmark$ & 624 (52\%) & 400\\
EyeQ \cite{fu2019evaluation} & 5 & 12,553 & $\checkmark$ & 2,330 (19\%) & 13,029\\
DRAC \cite{qian2024drac} & 3 & 611 & $\checkmark$ & 50 (8\%) & 386\\
ODIR \cite{odir2019peking} & 7 & 6,259 & $\checkmark$ & $\checkmark$ & 900 \\
\bottomrule
\end{tabular}
}
\vspace{-2.0em}
\end{center}
\end{table}

\begin{table*}[t]
\begin{center}
\caption{Results of disease diagnosis  (Kappa \%) on DDR with different ratios of symmetric and asymmetric noise. The top-1 and top-2 results are marked in \textbf{bold} and \underline{underline}.}
\label{table:comparison}
\begin{tabular}{lccccccccccccccccccc}
\toprule
& & \multicolumn{10}{c}{Symmetric} & \multicolumn{8}{c}{Asymmetric}\\
\cmidrule(lr){3-12}
\cmidrule(lr){13-20}
Mis-L ratio & & \multicolumn{2}{l}{15\%} & \multicolumn{2}{l}{15\%} & \multicolumn{2}{l}{15\%} & \multicolumn{2}{l}{35\%} & \multicolumn{2}{l}{55\%} & 
\multicolumn{2}{l}{15\%} & \multicolumn{2}{l}{15\%} & \multicolumn{2}{l}{30\%} & \multicolumn{2}{l}{45\%} \\
\midrule
Mis-H ratio & & \multicolumn{2}{l}{20\%} & \multicolumn{2}{l}{50\%} & \multicolumn{2}{l}{80\%} & \multicolumn{2}{l}{20\%} & \multicolumn{2}{l}{20\%} & \multicolumn{2}{l}{30\%} & \multicolumn{2}{l}{50\%} & \multicolumn{2}{l}{20\%} & \multicolumn{2}{l}{20\%} \\
\midrule
\multirow{2}{*}{Standard CE} 
& Best & 88.3 & \hspace{-12pt} \multirow{2}{*}{\color{NavyBlue}{$\downarrow$12.2}} & 81.4 & \hspace{-12pt} \multirow{2}{*}{\color{NavyBlue}{$\downarrow$41.7}} & 72.4 & \hspace{-12pt} \multirow{2}{*}{\color{NavyBlue}{$\downarrow$56.5}} & 86.7 & \hspace{-12pt} \multirow{2}{*}{\color{NavyBlue}{$\downarrow$17.3}} & 87.1 & \hspace{-12pt} \multirow{2}{*}{\color{NavyBlue}{$\downarrow$26.4}} & 85.3 & \hspace{-12pt} \multirow{2}{*}{\color{NavyBlue}{$\downarrow$19.1}} & 79.3 & \hspace{-12pt} \multirow{2}{*}{\color{NavyBlue}{$\downarrow$27.3}} & 85.9 & \hspace{-12pt} \multirow{2}{*}{\color{NavyBlue}{$\downarrow$10.3}} & 85.6 & \hspace{-12pt} \multirow{2}{*}{\color{NavyBlue}{$\downarrow$15.2}}\\
& Last & 76.1 & & 39.7 & & 15.9 & & 69.4 & & 60.7 & & 66.2 & & 52.0 & & 75.6 & & 70.4\\
\midrule
\midrule
\multirow{2}{*}{DivideMix \cite{li2020dividemix}} 
& Best & \underline{88.7} & \hspace{-12pt} \multirow{2}{*}{\color{NavyBlue}{$\downarrow$~1.5}} & 83.6  & \hspace{-12pt} \multirow{2}{*}{\color{NavyBlue}{$\downarrow$12.5}} & 71.6 & \hspace{-12pt} \multirow{2}{*}{\color{NavyBlue}{$\downarrow$28.7}} & 86.5 & \hspace{-12pt} \multirow{2}{*}{\color{NavyBlue}{$\downarrow$~1.7}} & 86.2 & \hspace{-12pt} \multirow{2}{*}{\color{NavyBlue}{$\downarrow$~3.7}} & 85.3 & \hspace{-12pt} \multirow{2}{*}{\color{NavyBlue}{$\downarrow$~1.3}} & 79.4 & \hspace{-12pt} \multirow{2}{*}{\color{NavyBlue}{$\downarrow$11.6}} & \underline{87.3} & \hspace{-12pt} \multirow{2}{*}{\color{NavyBlue}{$\downarrow$~2.7}} & 86.3 & \hspace{-12pt} \multirow{2}{*}{\color{NavyBlue}{$\downarrow$~2.0}} \\
& Last & 87.2 & & 71.1 & & 42.9 & & 84.8 & & 82.5 & & 84.0 & & 67.8 & & 84.6 & & \underline{84.3} \\
\midrule
\multirow{2}{*}{MOIT \cite{ortego2021multi}} 
& Best & 88.4 & \hspace{-12pt} \multirow{2}{*}{\color{NavyBlue}{$\downarrow$~3.6}} & 83.5 & \hspace{-12pt} \multirow{2}{*}{\color{NavyBlue}{$\downarrow$~0.0}} & 73.2 & \hspace{-12pt} \multirow{2}{*}{\color{NavyBlue}{$\downarrow$~8.0}} & 87.3 & \hspace{-12pt} \multirow{2}{*}{\color{NavyBlue}{$\downarrow$10.0}} & \underline{87.7} & \hspace{-12pt} \multirow{2}{*}{\color{NavyBlue}{$\downarrow$16.8}} & 85.4 & \hspace{-12pt} \multirow{2}{*}{\color{NavyBlue}{$\downarrow$~2.0}} & 82.0 & \hspace{-12pt} \multirow{2}{*}{\color{NavyBlue}{$\downarrow$~2.6}} & 86.7 & \hspace{-12pt} \multirow{2}{*}{\color{NavyBlue}{$\downarrow$~2.0}} & 85.6 & \hspace{-12pt} \multirow{2}{*}{\color{NavyBlue}{$\downarrow$~2.9}}\\
& Last & 84.8 & & \underline{83.5} & & 65.2 & & 77.3 & & 70.9 & & 83.4 & & \underline{79.4} & & \underline{84.7} & & 82.7\\
\midrule
\multirow{2}{*}{AugDMix \cite{nishi2021augmentation}} 
& Best & \underline{88.7}  & \hspace{-12pt} \multirow{2}{*}{\color{NavyBlue}{$\downarrow$~0.3}}& 83.1 & \hspace{-12pt} \multirow{2}{*}{\color{NavyBlue}{$\downarrow$~7.9}} & \underline{79.3} & \hspace{-12pt} \multirow{2}{*}{\color{NavyBlue}{$\downarrow$12.5}} & \underline{87.5} & \hspace{-12pt} \multirow{2}{*}{\color{NavyBlue}{$\downarrow$~2.6}} & 83.5 & \hspace{-12pt} \multirow{2}{*}{\color{NavyBlue}{$\downarrow$~4.7}} & \underline{86.7} & \hspace{-12pt} \multirow{2}{*}{\color{NavyBlue}{$\downarrow$~1.1}} & 81.8 & \hspace{-12pt} \multirow{2}{*}{\color{NavyBlue}{$\downarrow$15.0}} & 86.1 & \hspace{-12pt} \multirow{2}{*}{\color{NavyBlue}{$\downarrow$~2.6}} & 85.0 & \hspace{-12pt} \multirow{2}{*}{\color{NavyBlue}{$\downarrow$~2.6}}\\
& Last & \textbf{88.4} & & 75.2 & & 66.8 & & 84.9 & & 78.8 & & \underline{85.6} & & 66.8 & & 83.5 & & 82.4\\
\midrule
\multirow{2}{*}{UNICON \cite{karim2022unicon}} 
& Best & 87.9 & \hspace{-12pt} \multirow{2}{*}{\color{NavyBlue}{$\downarrow$~1.1}} & 83.5 & \hspace{-12pt} \multirow{2}{*}{\color{NavyBlue}{$\downarrow$12.6}} & 66.5 & \hspace{-12pt} \multirow{2}{*}{\color{NavyBlue}{$\downarrow$26.7}} & 86.7 & \hspace{-12pt} \multirow{2}{*}{\color{NavyBlue}{$\downarrow$~2.6}} & 86.8 & \hspace{-12pt} \multirow{2}{*}{\color{NavyBlue}{$\downarrow$~3.7}} & 86.6 & \hspace{-12pt} \multirow{2}{*}{\color{NavyBlue}{$\downarrow$~3.6}} & \underline{84.1} & \hspace{-12pt} \multirow{2}{*}{\color{NavyBlue}{$\downarrow$~4.7}} & 87.1 & \hspace{-12pt} \multirow{2}{*}{\color{NavyBlue}{$\downarrow$~4.1}} & \underline{86.7} & \hspace{-12pt} \multirow{2}{*}{\color{NavyBlue}{$\downarrow$~8.1}}\\
& Last & 86.8 & & 70.9 & & 39.8 & & 84.1 & & 83.1 & & 83.0 & & \underline{79.4} & & 83.0 &  & 78.6\\
\midrule
\multirow{2}{*}{PLReMix \cite{liu2024plremix}}  & Best & 86.4 & \hspace{-12pt} \multirow{2}{*}{\color{NavyBlue}{$\downarrow$~2.1}} & \underline{84.5} & \hspace{-12pt} \multirow{2}{*}{\color{NavyBlue}{$\downarrow$~2.8}} & 69.3 & \hspace{-12pt} \multirow{2}{*}{\color{NavyBlue}{$\downarrow$~8.5}} & 86.6  & \hspace{-12pt} \multirow{2}{*}{\color{NavyBlue}{$\downarrow$~3.7}} & 86.5 & \hspace{-12pt} \multirow{2}{*}{\color{NavyBlue}{$\downarrow$~4.2}} & 85.9 & \hspace{-12pt} \multirow{2}{*}{\color{NavyBlue}{$\downarrow$~3.1}} & 83.1 & \hspace{-12pt} \multirow{2}{*}{\color{NavyBlue}{$\downarrow$~4.8}} & 85.8 & \hspace{-12pt} \multirow{2}{*}{\color{NavyBlue}{$\downarrow$~5.5}} & 85.0 & \hspace{-12pt} \multirow{2}{*}{\color{NavyBlue}{$\downarrow$~8.1}}\\
& Last & 84.3 & & 81.7 & & 60.8 & & 82.9 & & 82.3 & & 82.8 & & 78.3 & & 80.3 & & 76.9\\
\midrule
\midrule
\multirow{2}{*}{SEAL-IDN \cite{chen2021beyond}} 
& Best & 87.3 & \hspace{-12pt} \multirow{2}{*}{\color{NavyBlue}{$\downarrow$~1.0}} & 83.3 & \hspace{-12pt} \multirow{2}{*}{\color{NavyBlue}{$\downarrow$~2.2}} & 75.3 & \hspace{-12pt} \multirow{2}{*}{\color{NavyBlue}{$\downarrow$~5.6}} & 86.7 & \hspace{-12pt} \multirow{2}{*}{\color{NavyBlue}{$\downarrow$~0.5}} & 87.3 & \hspace{-12pt} \multirow{2}{*}{\color{NavyBlue}{$\downarrow$~1.8}} & 84.1 & \hspace{-12pt} \multirow{2}{*}{\color{NavyBlue}{$\downarrow$~2.5}} & 78.2 & \hspace{-12pt} \multirow{2}{*}{\color{NavyBlue}{$\downarrow$~5.4}} & 85.8 & \hspace{-12pt} \multirow{2}{*}{\color{NavyBlue}{$\downarrow$~2.4}} & 84.9 & \hspace{-12pt} \multirow{2}{*}{\color{NavyBlue}{$\downarrow$~2.0}}\\
& Last & 86.3 & & 81.1 & & \underline{69.7} & & \underline{86.2} & & \underline{85.5} & & 81.6 & & 72.8 & & 83.4 & & 82.9\\
\midrule
\multirow{2}{*}{CD-CCR \cite{cheng2022class}} 
& Best & 87.3 & \hspace{-12pt} \multirow{2}{*}{\color{NavyBlue}{$\downarrow$~3.1}} & 82.6 & \hspace{-12pt} \multirow{2}{*}{\color{NavyBlue}{$\downarrow$21.3}} & 69.6 & \hspace{-12pt} \multirow{2}{*}{\color{NavyBlue}{$\downarrow$19.6}} & 86.8 & \hspace{-12pt} \multirow{2}{*}{\color{NavyBlue}{$\downarrow$~6.1}} & 86.7 & \hspace{-12pt} \multirow{2}{*}{\color{NavyBlue}{$\downarrow$12.0}} & 86.4 & \hspace{-12pt} \multirow{2}{*}{\color{NavyBlue}{$\downarrow$10.4}} & 78.6 & \hspace{-12pt} \multirow{2}{*}{\color{NavyBlue}{$\downarrow$19.8}} & 86.8 & \hspace{-12pt} \multirow{2}{*}{\color{NavyBlue}{$\downarrow$~8.5}} & 86.4 & \hspace{-12pt} \multirow{2}{*}{\color{NavyBlue}{$\downarrow$~9.9}}\\
& Last & 84.2 & & 61.3 & & 50.0 & & 80.7 & & 74.7 & & 76.0 & & 58.8 & & 78.3 & & 76.5\\
\midrule
\multirow{2}{*}{MEIDTM \cite{cheng2022instance}} 
& Best & 86.6 & \hspace{-12pt} \multirow{2}{*}{\color{NavyBlue}{$\downarrow$~5.9}} & 83.0 & \hspace{-12pt} \multirow{2}{*}{\color{NavyBlue}{$\downarrow$28.6}} & 72.8 & \hspace{-12pt} \multirow{2}{*}{\color{NavyBlue}{$\downarrow$34.5}} & 86.6 & \hspace{-12pt} \multirow{2}{*}{\color{NavyBlue}{$\downarrow$12.3}} & 85.5 & \hspace{-12pt} \multirow{2}{*}{\color{NavyBlue}{$\downarrow$21.8}} & 84.9 & \hspace{-12pt} \multirow{2}{*}{\color{NavyBlue}{$\downarrow$12.7}} & 81.3 & \hspace{-12pt} \multirow{2}{*}{\color{NavyBlue}{$\downarrow$23.2}} & 85.8 & \hspace{-12pt} \multirow{2}{*}{\color{NavyBlue}{$\downarrow$~6.8}} & 85.6 & \hspace{-12pt} \multirow{2}{*}{\color{NavyBlue}{$\downarrow$~8.4}}\\
& Last & 80.7 & & 54.4 & & 38.3 & & 74.3 & & 63.7 & & 72.2 & & 58.1 & & 79.0 & & 77.2\\
\midrule
\midrule
\multirow{2}{*}{QMix (ours)} 
& Best & \textbf{90.1} & \hspace{-12pt} \multirow{2}{*}{\color{NavyBlue}{$\downarrow$~1.9}} & \textbf{87.9} & \hspace{-12pt} \multirow{2}{*}{\color{NavyBlue}{$\downarrow$~1.6}} & \textbf{81.2} & \hspace{-12pt} \multirow{2}{*}{\color{NavyBlue}{$\downarrow$~5.3}} & \textbf{89.2} & \hspace{-12pt} \multirow{2}{*}{\color{NavyBlue}{$\downarrow$~2.1}} & \textbf{88.5} & \hspace{-12pt} \multirow{2}{*}{\color{NavyBlue}{$\downarrow$~1.3}} & \textbf{89.7} & \hspace{-12pt} \multirow{2}{*}{\color{NavyBlue}{$\downarrow$~0.8}} & \textbf{87.4} & \hspace{-12pt} \multirow{2}{*}{\color{NavyBlue}{$\downarrow$~2.1}} & \textbf{89.1} & \hspace{-12pt} \multirow{2}{*}{\color{NavyBlue}{$\downarrow$~1.7}} & \textbf{88.3} & \hspace{-12pt} \multirow{2}{*}{\color{NavyBlue}{$\downarrow$~1.6}} \\
& Last & \underline{88.2} & & \textbf{86.3} & & \textbf{75.9} & & \textbf{87.1} & & \textbf{87.2} & & \textbf{88.9} & & \textbf{85.3} & & \textbf{87.4} & & \textbf{86.7}\\
\bottomrule
\end{tabular}
\vspace{-2.0em}
\end{center}
\end{table*}

\subsection{Datasets}
\label{subsec:dataset}
We validate our method on five datasets for diabetic retinopathy (DR) grading and one dataset for recognition of multiple ocular diseases. The severity of DR can be divided into five stages \cite{wilkinson2003proposed}, including no DR, Mild Nonproliferative DR (NPDR), Moderate NPDR, Severe NPDR, and Proliferative DR (PDR). These five grades can also be merged into a ternary classification scheme \cite{qian2024drac}, consisting of no DR, NPDR, and PDR. 
The statistical details of these datasets are provided in Table \ref{table:dataset}.

For the synthetic noisy dataset DDR, it originally contains 6,260 Correct training samples and 1,151 Mis-L samples. 
Mis-H samples are generated by changing the labels of a portion of Correct samples in a symmetric or asymmetric manner. 
In specific, symmetric noise refers to generating the wrong labels under a uniform distribution across all five classes. 
Asymmetric noise indicates that assigning wrong labels based on the imbalanced categorical distribution of the dataset, posing additional challenges.
Additionally, it can be characterized by the confusion between similar categories, but determining the specific transition probabilities between different disease statuses is difficult. In this study, we adopt asymmetric noise based on the categorical distribution.
We experiment with different noise ratios by changing the number of Mis-H and Mis-L for a comprehensive comparison.
Notably, in addition to the provided Mis-L, we adopt an off-the-shelf low-quality retinal image generation algorithm \cite{deep_reitna_enhance} to create more Mis-L for higher rates.

In the group of real-world datasets, DRTiD \cite{hou2022cross} and DeepDRiD \cite{deepdrid} provide two-field color fundus photography (CFP), which capture the macula-centric and optic disc-centric views to enable a more comprehensive diagnosis. EyeQ \cite{fu2019evaluation} is a large-scale CFP dataset for five-level DR grading, while DRAC \cite{qian2024drac} is an optical coherence tomography angiography (OCTA) dataset for ternary DR screening. 
The original ODIR dataset is a multi-label dataset consisting of 8 classes, including normal, diabetes, glaucoma, cataract, age-related macular degeneration (AMD), hypertension, myopia and other diseases/abnormalities, with 7,000 training and 1,000 off-site testing samples. However, since our method and all other comparable methods are intended for single-label classification, we exclude images with multi-label annotations. Consequently, we also remove the AMD category, as all its images are multi-labeled.
Real-world retinal image datasets are prone to containing mislabeled images with both high and low quality, due to variations in imaging conditions, differences in equipment, and doctors' experience.
\#Mis-L is available as these real-world noisy datasets provide image quality labels.
DRTiD provides (1) initial disease labels from community screening that are not reviewed by ophthalmologists; and (2) image quality labels and disease labels verified by three experienced ophthalmologists, serving as the ground-truth. 
We consider images with original labels that are inconsistent with verified ground-truth labels and have high image quality as Mis-H.
In test sets, Mis-L samples are excluded.

\subsection{Experimental Setups}\label{subsec:implement}

\subsubsection{Implementation details}
Our model consists of a ResNet-50~\cite{he2016resnet} encoder pre-trained on ImageNet~\cite{deng2009imagenet} and a two-layer MLP as the projection network. 
Input images are resized to $384\times 384$. In sample separation, no data augmentation is applied to input images.
In quality-aware SSL training, weak augmentations include random horizontal flip, vertical flip and rotation. Strong augmentations include random horizontal flip, vertical flip, color jittering, random cropping, affine transformation, grayscaling, etc. 
We also adopt the MixUp \cite{zhang2018mixup} strategy.
To alleviate the issue of data imbalance, we conduct sample separation within each category's samples rather than across the entire training set.
The network is trained for 100 epochs with a batch size of 16 and a learning rate of 1e-3. 
To ensure reliable sample separation, the network first undergoes warm-up training for 10 epochs using the standard CE loss.
After that, we alternate sample separation and SSL training in each epoch. We set $\tau=0.07$, $\lambda_m=1$, and specific values of other hyperparameters are available in the implementation.
We employ Quadratic Weighted Kappa \cite{kaggle} as the evaluation metric for disease diagnosis and AUC score for sample separation.
To show the robustness of LNL methods, we report both the model checkpoints with the best result across all epochs (Best) and the average of the last 10 epochs (Last). 
If the model suffers noisy samples, the Last checkpoint is expected to be significantly lower than Best.

\subsubsection{Compared methods}
The baseline method is a disease diagnosis model trained by the standard CE loss, without considering the noisy images in the datasets.
We compare QMix with other state-of-the-art LNL methods. DivideMix \cite{li2020dividemix}, MOIT \cite{ortego2021multi}, AugDMix \cite{nishi2021augmentation}, UNICON \cite{karim2022unicon}, PLReMix \cite{liu2024plremix} are sample selection based methods with SSL training. 
SEAL-IDN \cite{chen2021beyond}, CD-CCR \cite{cheng2022class}, MEIDTM \cite{cheng2022instance} are particularly proposed for instance-dependent label noise, where mislabeling is dependent on the specific visual appearance of each image, \emph{e.g.}, visual ambiguity. Low-quality images can also be viewed as a source of instance-dependent label noise \cite{cheng2022instance}.

\subsection{Results on Synthetic Noisy Datasets}
\label{subsec:synthetic}

First, we evaluate our method on the synthetic noisy label dataset DDR for five-level DR grading.
Table \ref{table:comparison} shows the disease diagnosis results on DDR with different levels of symmetric and asymmetric label noise.
With the increase of noise ratios of Mis-L and Mis-H, respectively, QMix consistently outperforms all other state-of-the-art LNL methods, except for a Last model under a low noise rate (15\% Mis-L and 20\% Mis-H).
In the case of symmetric noise, for high noise rates (\emph{e.g.}, 80\% Mis-H ratio), the Last checkpoint of the baseline CE method experiences a significant performance drop (72.4\% $\rightarrow$ 15.9\%). This indicates that the noisy labels over-fitted by the baseline model greatly affect disease diagnosis performance. 
Other LNL approaches, \emph{e.g.}, DivideMix, UNICON, and MEIDTM, also suffer from the high noise ratio, exhibiting around a 30\% performance drop. Similar results can be observed under a high Mis-L noise ratio of 55\%. 
In comparison, QMix achieves substantial performance improvements over these state-of-the-arts.
In the case of asymmetric noise, both correct and mislabeled samples share an imbalanced categorical distribution. This makes the selection of correct samples more challenging. However, the improvements of our QMix are significant under asymmetric noise across all noise rates. For instance, under a 45\% Mis-L and 20\% Mis-H noise ratio, QMix achieves 88.3\% and 86.7\% for the Best and Last checkpoints, respectively. This indicates that our model is robust to these mislabeled images during the entire training
process.

\begin{table}
\begin{center}
\caption{Results of disease diagnosis (Kappa \%) on five real-world noisy retinal image datasets.}
\label{table:real-world}
\resizebox{\columnwidth}{!}{
\begin{tabular}{lccccccccc}
\toprule
Method & & DRTiD & DeepDRiD & EyeQ & DRAC & ODIR \\
\midrule
\multirow{2}{*}{Standard CE} 
& Best & 72.0 & 91.7 & 78.4 & 82.5 & 56.7\\
& Last & 51.3 & 87.8 & 75.7 & 74.1 & 54.6\\
\midrule
\midrule
\multirow{2}{*}{DivideMix \cite{li2020dividemix}}
& Best & \underline{78.3} & 93.0 & 76.3 & 82.7 & 57.7\\
& Last & 74.9 & \underline{91.7} & 73.8 & 80.2 & \underline{56.7}\\
\midrule
\multirow{2}{*}{MOIT \cite{ortego2021multi}}
& Best & 74.1 & \textbf{94.1} & \underline{80.7} & \underline{83.5} & 56.2 \\
& Last & 63.1 & 90.0 & 77.2 & 79.6 & 49.5\\
\midrule
\multirow{2}{*}{AugDMix \cite{nishi2021augmentation}}
& Best & 77.6 & 92.6 & 76.6 & 81.2 & 57.2 \\
& Last & \underline{75.4} & 88.2 & 69.0 & 78.3 & 52.9\\
\midrule
\multirow{2}{*}{UNICON \cite{karim2022unicon}}
& Best & 77.9 & 92.3 & 78.1 & 82.7 & 55.0 \\
& Last & 75.1 & 88.3 & 68.4 & 76.0 & 48.5 \\
\midrule
\multirow{2}{*}{PLReMix \cite{liu2024plremix}} 
& Best & 76.7 & 91.8 & 76.2 & 81.6 & 46.4 \\
& Last & 69.4 & 84.2 & 51.3 & 76.6 & 42.8 \\
\midrule
\midrule
\multirow{2}{*}{SEAL-IDN \cite{chen2021beyond}}
& Best & 75.4 & \underline{93.6} & 78.8 & 83.3 & \underline{58.5} \\
& Last & 70.2 & 88.1 & 76.3 & 79.6 & 53.1 \\
\midrule
\multirow{2}{*}{CD-CCR \cite{cheng2022class}} 
& Best & 73.4 & 91.5 & 79.2 & 83.4 & 54.5\\
& Last & 55.9 & 89.0 & \underline{78.1} & \underline{81.5} & 52.6 \\
\midrule
\multirow{2}{*}{MEIDTM \cite{cheng2022instance}}
& Best & 74.7 & 91.2 & 78.9 & 82.4 & 56.6 \\
& Last & 60.9 & 88.2 & 77.1 & 78.1 & 54.3 \\
\midrule
\midrule
\multirow{2}{*}{QMix (ours)} 
& Best & \textbf{80.8} & \textbf{94.1} & \textbf{80.9} & \textbf{84.1} & \textbf{61.4} \\
& Last & \textbf{78.4} & \textbf{92.2} & \textbf{79.1} & \textbf{81.7} & \textbf{57.2} \\
\bottomrule
\end{tabular}
}
\end{center}
\end{table}

\subsection{Results on Real-world Noisy Datasets} 
\label{subsec:realworld}

Table \ref{table:real-world} illustrates the superior disease diagnosis performance of our proposed QMix when training in the presence of real-world label noise on five retinal image datasets.

\noindent\textbf{DRTiD.}
When trained by the standard CE loss, the network shows a more than 20\% performance decrease from the Best checkpoint to the Last. 
Similarly, the performance of other LNL methods such as CD-CCR and MEIDTM drops by a large margin (17.5\% and 13.8\%, respectively). This reveals that these methods are less robust when dealing with mixed noise (31\% Mis-H and 26\% Mis-L). 
In contrast, QMix achieves significant improvements over the second-best performing methods, \emph{i.e.}, DivideMix and AugDMix, on the results of Best and Last checkpoints (2.5\% and 3.0\%, respectively).

\noindent\textbf{DeepDRiD.}
Results on DeepDRiD further confirm the substantial improvements achieved by our model for retinal disease diagnosis using two-field fundus images. Despite the high Mis-L ratio of 52\% in DeepDRiD, QMix retains its performance and demonstrates robustness to over-fitting on noisy samples when the training set is comparably small, containing only 1,200 images. 

\noindent\textbf{EyeQ.}
Though EyeQ contains a relatively low proportion (18.5\%) of Mis-L, our QMix still achieves an approximately 2\% improvement on both the Best and Last checkpoints compared to the baseline Standard CE method. 
Furthermore, when compared to the other LNL methods, we achieve a superior Best result of 80.9\%, while also obtaining a higher performance of 79.1\$ on the Last checkpoint. The proposed QMix demonstrates robustness on EyeQ.

\noindent\textbf{DRAC.}
On this OCTA dataset, the baseline Standard CE method also suffers from the over-fitting problem (82.5\%$\rightarrow$74.1\%).
In the face of the challenge posed by noisy samples, QMix achieves the best performance with a Kappa of 84.1\% on the Best checkpoint and 81.7\% on the Last checkpoint, showcasing its robustness to noise on different modalities of retinal images.

\noindent\textbf{ODIR.}
Compared to other methods, our QMix demonstrates superior and more robust performance in the more challenging task of multiple ocular disease recognition. It achieved 3$\sim$4\% improvements in the Kappa score over the baseline standard CE. This enhancement underscores the effectiveness of QMix in classifying a variety of ocular diseases in real-world situations that involve mixed noise.

\subsection{Sample Separation Performance}\label{subsec:sampleseparation}

\begin{figure}[t]
\begin{center}
   \includegraphics[width=\columnwidth]{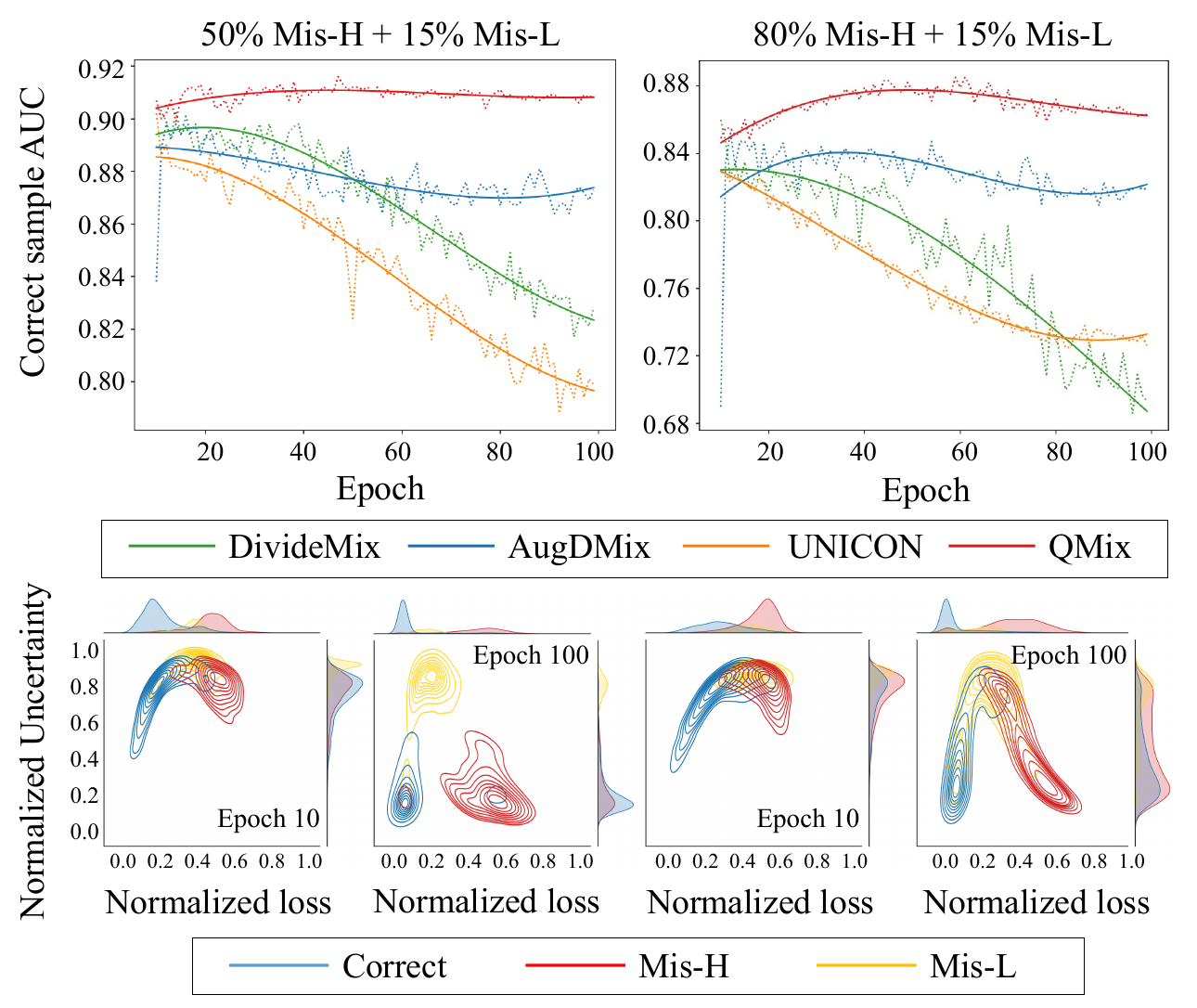}
\end{center}
\vspace{-1.0em}
   \caption{Sample separation performance under symmetric noise on the DDR dataset. Top: Comparison of correct sample AUC. Bottom: Visualization of the joint uncertainty-loss distribution. }
\label{fig:clean_auc}
\vspace{-1.0em}
\end{figure}

Accurate selection of correct samples is crucial for achieving high disease diagnosis performance. 
We compare the ROC-AUC scores between our sample separation mechanism and other methods \cite{li2020dividemix,nishi2021augmentation,karim2022unicon} on the DDR dataset. 
As illustrated in Fig. \ref{fig:clean_auc} top, the correct sample AUC scores of DivideMix and UNICON decline significantly as the training progresses. AugDMix also has a difficulty in distinguishing Correct in the presence of mixed noise, as these methods rely solely on the small-loss criterion to differentiate Mis-H from Correct.
In comparison, QMix maintains a high AUC of separating Correct throughout the entire training process, which benefits the quality-aware SSL training. 
Fig. \ref{fig:clean_auc} bottom shows the joint uncertainty-loss distribution of training samples under ``50\% Mis-H + 15\% Mis-L'' and ``80\% Mis-H + 15\% Mis-L'' noise rates. 
After 10 epochs of warm-up training, Correct is initially mixed with Mis-H and Mis-L.
During the quality-aware SSL training, our QMix manages to separate three types of training samples into distinct clusters by the final epoch.

\subsection{Ablation Study and Discussion}
\label{subsec:ablation}
We conduct ablation studies on DDR to evaluate the effects of
(1) network co-training, ensemble, and MixUp;
(2) joint uncertainty-loss criterion; 
(3) sample-reweighing loss and contrastive enhancement loss in quality-aware SSL training.
Table \ref{table:ablation_sum} shows the results of disease diagnosis and Mis-L sample separation.
We also analyze the impact of different labeling weights for Correct and Mis-L in Table \ref{table:ablation_weight}. Finally, we visualize training sample distributions in Fig. \ref{fig:umap} and examples identified as Mis-L and Mis-H by QMix in Fig. \ref{fig:misl} and Fig. \ref{fig:mish}, respectively.

\begin{table}[t]
\begin{center}
\caption{Results of disease diagnosis and Mis-L sample separation under different types and levels of noise on DDR. }
\label{table:ablation_sum}
\begin{tabular}{lccccc}
\toprule
& & \multicolumn{2}{c}{Symmetric} & \multicolumn{2}{c}{Asymmetric}\\
\cmidrule(lr){3-4}
\cmidrule(lr){5-6}
Mis-L ratio & & 15\% & 15\% & 15\% & 15\% \\
\midrule
Mis-H ratio & & 50\% & 80\% & 30\% & 50\% \\
\midrule
\rowcolor{mygray}\multicolumn{6}{l}{Disease Diagnosis (Kappa \%)}\\
\midrule
\multirow{2}{*}{1~~~QMix} 
& Best & \textbf{87.9} & \textbf{81.2} & \textbf{89.7} & \textbf{87.4} \\
& Last & \textbf{86.3} & \textbf{75.9} & \textbf{88.9} & \textbf{85.3} \\
\midrule
\multirow{2}{*}{2~~~QMix w/o ensemble} 
& Best & 86.9 & 78.6 & 89.2 & 86.3 \\
& Last  & 85.4 & 74.0 & 88.7 & 84.0 \\
\midrule
\multirow{2}{*}{3~~~QMix w/o co-training} 
& Best  & 87.0 & 73.4 & 89.0 & 85.7 \\
& Last & 84.4 & 32.3 & 87.6 & 83.7 \\
\midrule
\multirow{2}{*}{4~~~QMix w/o MixUp} 
& Best & 85.9 & 77.7 & 89.6 & 85.9\\
& Last & 80.5 & 61.5 & 84.5 & 75.9\\
\midrule
\multirow{2}{*}{5~~~QMix w/o uncertainty}
& Best & 87.3 & 74.6 & 89.1 & 87.0 \\
& Last & 83.8 & 47.3 & 83.7 & 80.2 \\
\midrule
\multirow{2}{*}{6~~~QMix w/o $\mathcal{L}_{mis-l}$} 
& Best & 87.6 & 71.0 & 87.2 & 86.7 \\
& Last & 84.7 & 65.2 & 84.8 & 82.8 \\
\midrule
\multirow{2}{*}{7~~~QMix w/o $\mathcal{L}_{con}$} 
& Best & 86.6 & 78.2 & 89.1 & 85.5 \\
& Last & 83.9 & 69.5 & 87.2 & 83.6 \\
\midrule
\rowcolor{mygray}\multicolumn{6}{l}{Mis-L Sample Separation (AUC \%)}\\
\midrule
\multirow{2}{*}{8~~~QMix} 
& Best & \textbf{98.1} & \textbf{91.6} & \textbf{92.6} & \textbf{86.2} \\
& Last & \textbf{96.1} & \textbf{69.3} & \textbf{89.1} & \textbf{80.3} \\
\midrule
\multirow{2}{*}{9~~~QMix w/o uncertainty}
& Best & 81.9 & 76.7 & 81.7 & 74.6 \\
& Last & 63.6 & 53.3 & 58.2 & 52.8 \\
\midrule
\multirow{2}{*}{10~~~QMix w/o $\mathcal{L}_{mis-l}$} 
& Best & 81.6 & 63.5 & 87.6 & 81.6 \\
& Last & 62.1 & 55.3 & 76.4 & 78.1 \\
\midrule
\multirow{2}{*}{11~~QMix w/o $\mathcal{L}_{con}$} 
& Best & 97.6 & 91.4 & 90.7 & 76.1 \\
& Last & 92.0 & 63.7 & 87.1 & 65.8 \\
\bottomrule
\end{tabular}
\vspace{-2.0em}
\end{center}
\end{table}

\subsubsection{Effect of network co-training, ensemble, and MixUp}
Network co-training, ensemble, and MixUp are common strategies in LNL methods \cite{li2020dividemix,nishi2021augmentation,karim2022unicon,liu2024plremix}. Results from the 2nd and 4th rows of Table \ref{table:ablation_sum} demonstrate the effectiveness of the three strategies in our proposed QMix. Co-training proved to be significant to avoid error accumulation particularly when dealing with high noise rates, \emph{e.g.}, 80\% Mis-H ratio.

\subsubsection{Effect of joint uncertainty-loss criterion} 
We compare our joint uncertainty-loss criterion with the basic small-loss criterion that only leverages the loss value for sample separation. 
For disease diagnosis results, as observed in Table \ref{table:ablation_sum}, ``QMix w/o uncertainty'' leads to a substantial performance drop on the Last checkpoint, especially for high noise rates, \emph{e.g.}, 80\% Mis-H ratio. 
In comparison, our QMix demonstrates superior and robust diagnosis results under both symmetric and asymmetric scenarios. The underlying reason for this is the improved sample separation. As shown in the 8th and 9th rows of Table \ref{table:ablation_sum}, QMix outperforms ``QMix w/o uncertainty'' by consistently achieving 10\%$\sim$30\% improvements on both the Best and Last checkpoints.

\subsubsection{Effect of sample-reweighing loss} 
We evaluate the effectiveness of the sample-reweighing loss $\mathcal{L}_{mis-l}$ on Mis-L. 
``QMix w/o $\mathcal{L}_{mis-l}$'' is achieved by setting the weight $\Omega(t)$ as a constant zero value, discarding the selected Mis-L during the entire training process. 
As revealed in the 6th row of Table \ref{table:ablation_sum}, the diagnosis performance on both the Best and Last checkpoints demonstrates a considerable decrease, especially at high noise ratios.
This is due to the insufficient model capability in sample separation during the early training epochs, where the selected Mis-L set might contains partial correct samples.
Discarding $\mathcal{L}_{mis-l}$ results in the loss of useful correct samples. 
In contrast, our sample-reweighing loss with a linear weight decay strategy begins with a very small weight and gradually decays to zero, preventing the loss of useful information in the early training epochs. 
It enhances the sample separation process (8th row vs. 10th row of Table \ref{table:ablation_sum}), and thus boosts diagnosis accuracy, particularly in the case of challenging asymmetric noise.

\subsubsection{Effect of contrastive enhancement loss}
We demonstrate the impact of the contrastive enhancement loss $\mathcal{L}_{con}$ in the 7th and 11th rows of Table \ref{table:ablation_sum}.
As observed, incorporating $\mathcal{L}_{con}$ proved to be effective for improving disease diagnosis and sample separation performance across all the noise scenarios. 
$\mathcal{L}_{con}$ pulls the representations of \emph{any disease category} in correct samples far from that of Mis-L in the latent space, while maintaining the distance between different  disease categories.
This allows the model to focus on disease-relevant features while minimizing the interference of Mis-L during training. By discriminating Mis-L and Correct, the sample separation process in the next round is expected to be more accurate.

\subsubsection{Analysis of labeling weights}
During the quality-aware SSL training process, Correct and Mis-H are treated as labeled and unlabeled data, respectively, while Mis-L is gradually discarded. 
Table \ref{table:ablation_weight} presents the results of disease diagnosis and sample separation under different choices of sample labeling in Eq. (\ref{eq:co-refinement}). The weights $w^0, w^1, w^2$ denote the probability of samples being classified by GMM as Correct, Mis-L, Mis-H in Eq. (\ref{eq:w}).
In the case where the standard co-refinement is used for Correct, using co-refinement for Mis-L (2nd row) achieves better results than co-guessing (1st row).
This is because co-refinement preserves the effective information (\emph{i.e.}, original labels) due to the network's confusion between Correct and Mis-L in the early training stage.
Therefore, our default choice (3rd row) sets the two co-refinement weights as $w^0+w^2$, leading to the best overall performance.

\begin{table}[t]
\begin{center}
\caption{Weights in the labeling branch (Eq. \ref{eq:co-refinement}) for predicted Correct and Mis-L under different noise rates on DDR. }
\label{table:ablation_weight}
\begin{tabular}{lcccccccccc}
\toprule
\multicolumn{2}{l}{Mis-L ratio} & 15\% & 15\% & 15\% & 15\%\\
\midrule
\multicolumn{2}{l}{Mis-H ratio} & 50\% & 80\% & 50\% & 80\%\\
\midrule
$(w^{correct}$, $w^{mis-l})$ & & \multicolumn{2}{c}{D. Diagnosis} & \multicolumn{2}{c}{S. Separation}\\
\midrule
\multirow{2}{*}{1~~~$(w^0, 0)$} & Best & 84.0 & 68.9 & 96.3 & 80.4\\
& Last & 83.1 & 60.0 & 91.9 & 67.4\\
\midrule
\multirow{2}{*}{2~~~$(w^0, w^2)$} & Best & 85.6 & 78.7 & 98.1 & 89.8\\
& Last & 84.8 & 74.8 & 94.8 & 68.3\\
\midrule
\multirow{2}{*}{3~~~$(w^0+w^2, w^0+w^2)$} & Best & \textbf{87.9} & \textbf{81.2} & \textbf{98.1}  & \textbf{91.6} \\
& Last & \textbf{86.3} & \textbf{75.9} & \textbf{96.1} & \textbf{69.3} \\
\bottomrule
\end{tabular}
\vspace{-1.0em}
\end{center}
\end{table}

\subsubsection{UMAP visualization}
\begin{figure}[t]
\begin{center}
   \includegraphics[width=\linewidth]{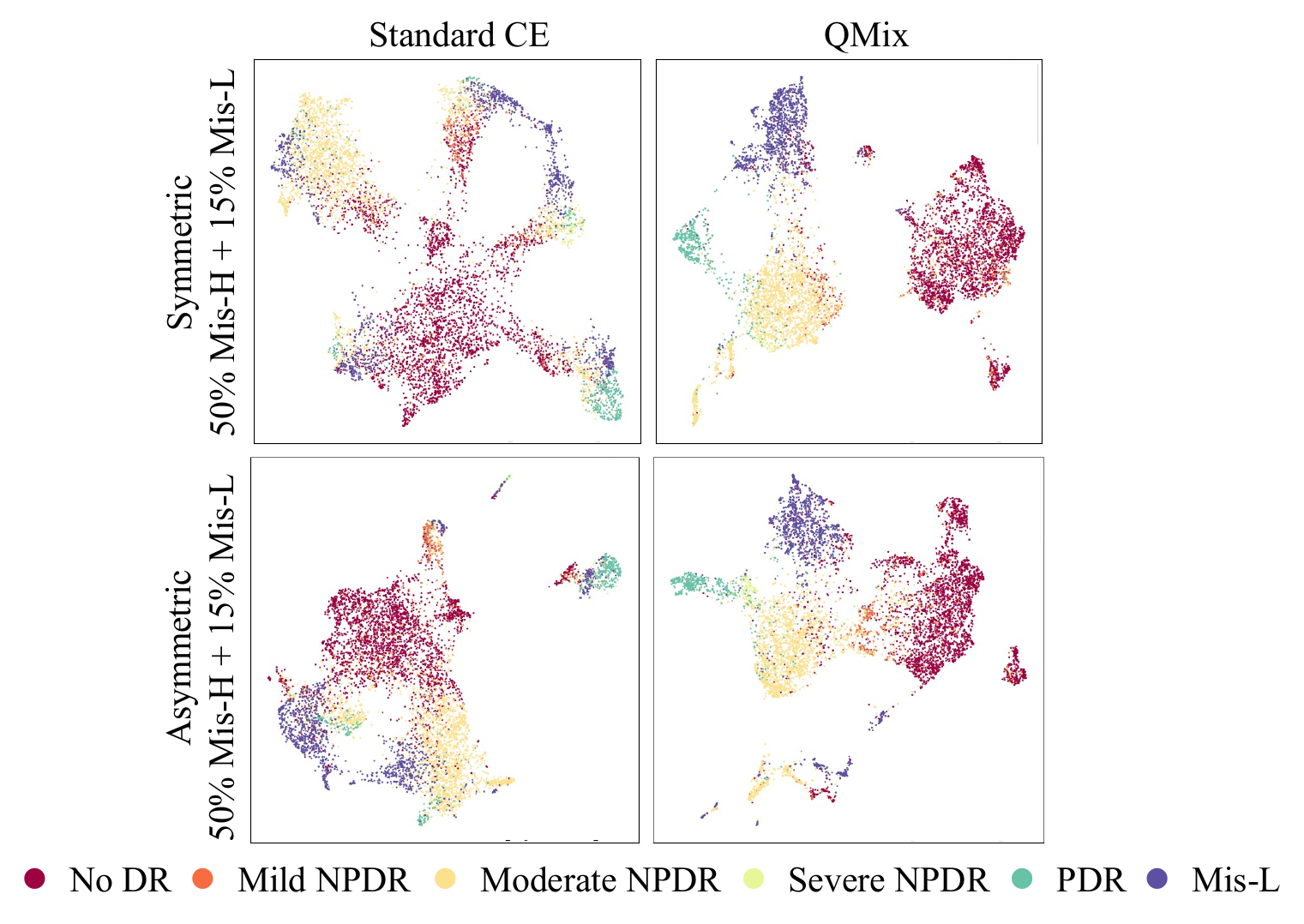}
\end{center}
\vspace{-1.0em}
   \caption{UMAP of DDR training images of Standard CE and QMix after 100 epochs of training under symmetric and asymmetric 50\% Mis-H + 15\% Mis-L noise ratios. Different colors indicate true labels.}
\label{fig:umap}
\vspace{-1.0em}
\end{figure}

We visualize the representations of training images using uniform manifold approximation and projection (UMAP) \cite{mcinnes2018umap}. We compare the standard CE and QMix under symmetric and asymmetric 50\% Mis-H + 15\% Mis-L noise ratios on DDR after 100 epochs of training. As illustrated in Fig. \ref{fig:umap}, CE fails to learn distinctive representations for retinal disease diagnosis under mixed noise. 
In contrast, the embeddings obtained by QMix form distinct clusters for different DR grades. Furthermore, Mis-L is successfully distinguished from the five categories of disease due to contrastive enhancement. This demonstrates the effectiveness of QMix in separating samples of different classes as well as Mis-L.
However, the clusters of mild NPDR and severe NPDR are not well separated. This is due to the fact that these two categories contain much fewer samples compared to the other classes, making it more challenging to learn distinctive features.

\subsubsection{Noisy samples in real-world dataset}
\begin{figure}[t]
\begin{center}
   \includegraphics[width=\linewidth]{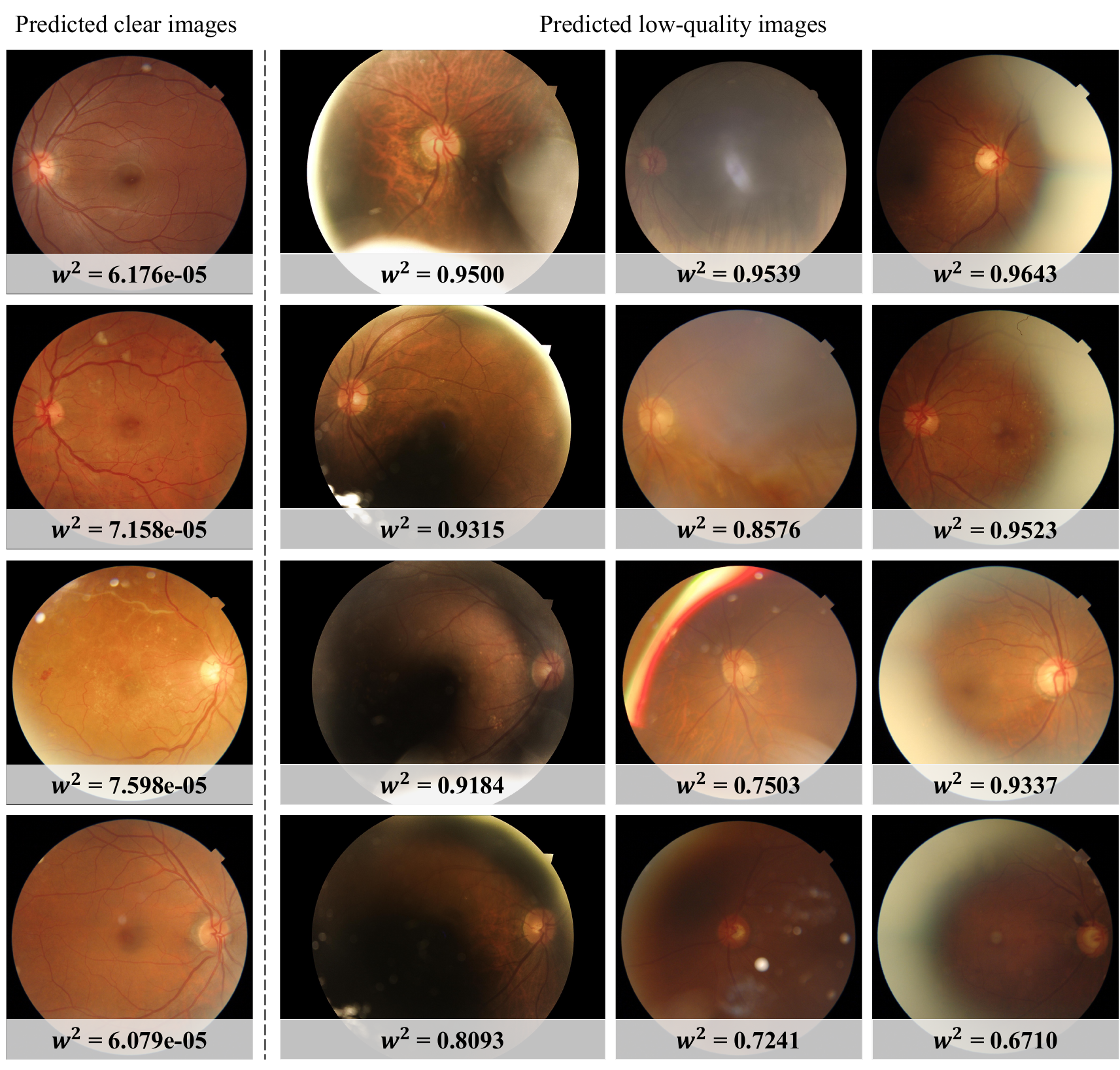}
\end{center}
\vspace{-1.0em}
   \caption{Examples of identified Mis-L by QMix on DRTiD. The numbers indicate the probability of the sample being Mis-L.}
\label{fig:misl}
\end{figure}

\begin{figure}[t]
\begin{center}
   \includegraphics[width=\linewidth]{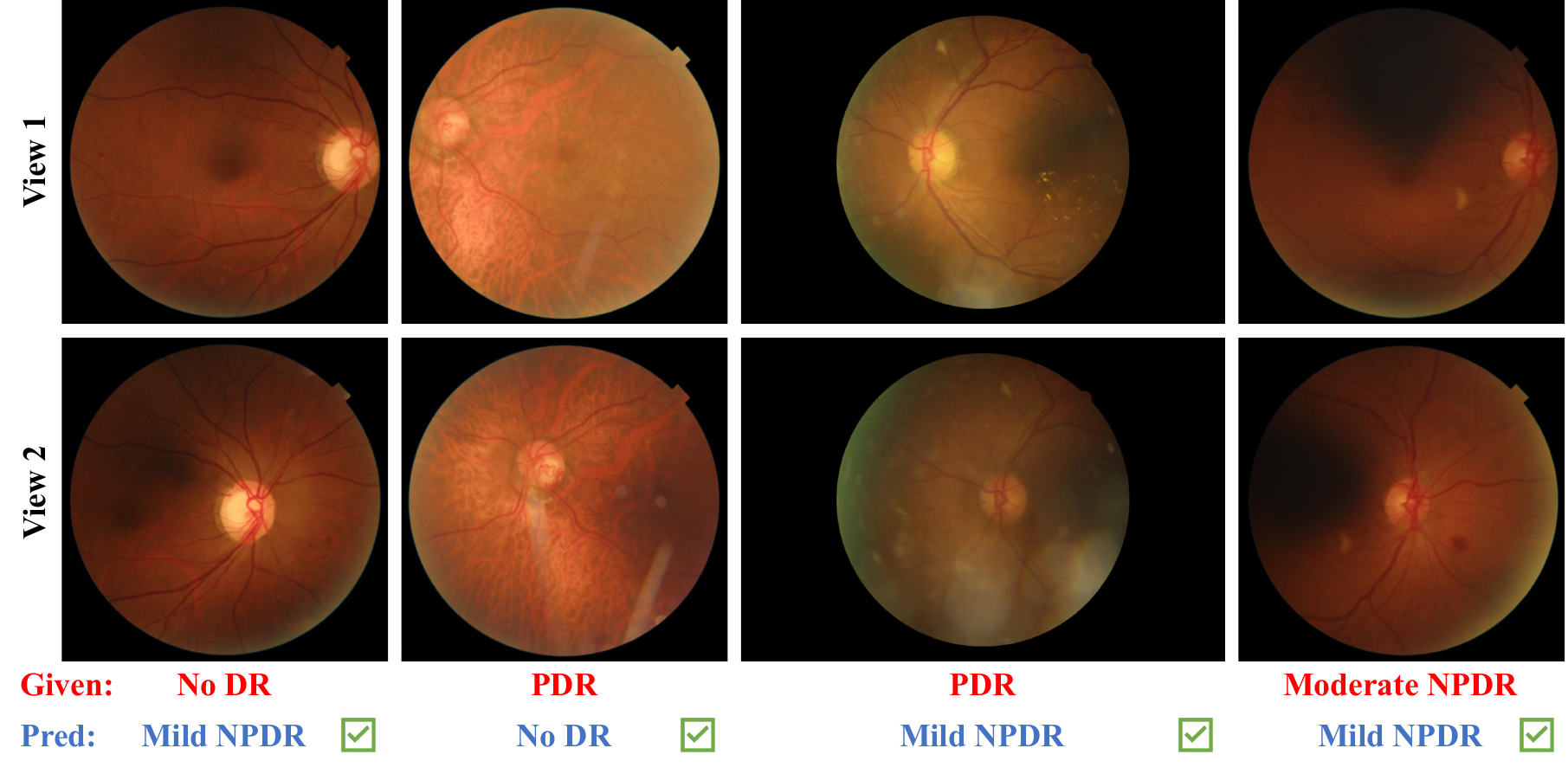}
\end{center}
\vspace{-1.0em}
   \caption{Two-field fundus images identified as Mis-H by QMix on DRTiD. 
   }
\label{fig:mish}
\vspace{-1.0em}
\end{figure}

We visualize some examples identified as Mis-L and Mis-H from the real-world DRTiD dataset \cite{hou2022cross}. 
In Fig. \ref{fig:misl}, we show the retinal images with their probabilities $w^2$ of being classified as Mis-L. 
In the 1st column, four clear retinal images are identified with very low probabilities of being Mis-H or Mis-L. 
In comparison, retinal images in the 2nd to 4th columns suffer from image quality issues such as artifacts, blur, and uneven illumination. 
These images are assigned with high Mis-L probabilities, which demonstrates the validity of our proposed sample separation criterion.
As shown in Fig. \ref{fig:mish}, we present several examples of two-field fundus images being classified as Mis-H from the DRTiD dataset. 
The information from two views is highly correlated and complementary, and benefits the comprehensive decision-making. Our QMix can effectively recognize the incorrect given labels and generate precise predictions, which are consistent with the true labels from ophthalmologists.

\section{Conclusion}
In this paper, we tackle the problem of learning a robust disease diagnosis model in the presence of mixed noise, where label noise and data noise exist in the dataset simultaneously.
We propose a novel LNL framework, termed as QMix, that alternates between sample separation and quality-aware SSL training in each epoch. 
In particular, through comprehensive studies on the memorization effect of modern DNNs under mixed noise, we design a joint uncertainty-loss criterion to effectively separate Correct/Mis-H/Mis-L images.
In the SSL training phase, we propose two novel loss functions, \emph{i.e.}, sample re-weighing loss and contrastive enhancement loss, to mitigate the negative impact of Mis-L on network training and enhance disease-specific feature learning. 
Extensive experiments on six public retinal image datasets demonstrated the superiority and robustness of QMix over state-of-the-art methods. 
However, when the characteristics of the retinal disease and low-quality samples are similar, such as both cataracts and low-quality samples exhibiting blurriness and unclear retinal structures, the sample separation may lead to confusion between the two. In future work, we plan to explore a more effective and precise sample separation strategy to deal with similar samples.

\bibliographystyle{IEEEtran}
\bibliography{refs}

\begin{thebibliography}{10}
\providecommand{\url}[1]{#1}
\csname url@samestyle\endcsname
\providecommand{\newblock}{\relax}
\providecommand{\bibinfo}[2]{#2}
\providecommand{\BIBentrySTDinterwordspacing}{\spaceskip=0pt\relax}
\providecommand{\BIBentryALTinterwordstretchfactor}{4}
\providecommand{\BIBentryALTinterwordspacing}{\spaceskip=\fontdimen2\font plus
\BIBentryALTinterwordstretchfactor\fontdimen3\font minus \fontdimen4\font\relax}
\providecommand{\BIBforeignlanguage}[2]{{%
\expandafter\ifx\csname l@#1\endcsname\relax
\typeout{** WARNING: IEEEtran.bst: No hyphenation pattern has been}%
\typeout{** loaded for the language `#1'. Using the pattern for}%
\typeout{** the default language instead.}%
\else
\language=\csname l@#1\endcsname
\fi
#2}}
\providecommand{\BIBdecl}{\relax}
\BIBdecl

\bibitem{karimi2020deep}
D.~Karimi, H.~Dou, S.~K. Warfield, and A.~Gholipour, ``Deep learning with noisy labels: Exploring techniques and remedies in medical image analysis,'' \emph{Medical Image Analysis}, vol.~65, p. 101759, 2020.

\bibitem{zhang2021understanding}
C.~Zhang, S.~Bengio, M.~Hardt, B.~Recht, and O.~Vinyals, ``Understanding deep learning (still) requires rethinking generalization,'' \emph{Communications of the ACM}, vol.~64, no.~3, pp. 107--115, 2021.

\bibitem{song2022learning}
H.~Song, M.~Kim, D.~Park, Y.~Shin, and J.-G. Lee, ``Learning from noisy labels with deep neural networks: A survey,'' \emph{IEEE Transactions on Neural Networks and Learning Systems}, 2022.

\bibitem{algan2021image}
G.~Algan and I.~Ulusoy, ``Image classification with deep learning in the presence of noisy labels: A survey,'' \emph{Knowledge-Based Systems}, vol. 215, p. 106771, 2021.

\bibitem{jiang2018mentornet}
L.~Jiang, Z.~Zhou, T.~Leung, L.-J. Li, and L.~Fei-Fei, ``Mentornet: Learning data-driven curriculum for very deep neural networks on corrupted labels,'' in \emph{International Conference on Machine Learning}.\hskip 1em plus 0.5em minus 0.4em\relax PMLR, 2018, pp. 2304--2313.

\bibitem{han2018coteaching}
B.~Han \emph{et~al.}, ``Co-teaching: Robust training of deep neural networks with extremely noisy labels,'' \emph{Advances in Neural Information Processing Systems}, vol.~31, 2018.

\bibitem{li2020dividemix}
J.~Li, R.~Socher, and S.~C. Hoi, ``Dividemix: Learning with noisy labels as semi-supervised learning,'' in \emph{International Conference on Learning Representations}, 2020.

\bibitem{guo2018curriculumnet}
S.~Guo \emph{et~al.}, ``Curriculumnet: Weakly supervised learning from large-scale web images,'' in \emph{Proceedings of the European Conference on Computer Vision}, 2018, pp. 135--150.

\bibitem{karim2022unicon}
N.~Karim, M.~N. Rizve, N.~Rahnavard, A.~Mian, and M.~Shah, ``Unicon: Combating label noise through uniform selection and contrastive learning,'' in \emph{Proceedings of the IEEE/CVF Conference on Computer Vision and Pattern Recognition}, 2022, pp. 9676--9686.

\bibitem{li2019diagnostic}
T.~Li, Y.~Gao, K.~Wang \emph{et~al.}, ``Diagnostic assessment of deep learning algorithms for diabetic retinopathy screening,'' \emph{Information Sciences}, vol. 501, pp. 511--522, 2019.

\bibitem{hou2022cross}
J.~Hou \emph{et~al.}, ``Cross-field transformer for diabetic retinopathy grading on two-field fundus images,'' in \emph{2022 IEEE International Conference on Bioinformatics and Biomedicine (BIBM)}.\hskip 1em plus 0.5em minus 0.4em\relax IEEE, 2022, pp. 985--990.

\bibitem{deepdrid}
R.~Liu \emph{et~al.}, ``Deepdrid: Diabetic retinopathy—grading and image quality estimation challenge,'' \emph{Patterns}, p. 100512, 2022.

\bibitem{fu2019evaluation}
H.~Fu \emph{et~al.}, ``Evaluation of retinal image quality assessment networks in different color-spaces,'' in \emph{International Conference on Medical Image Computing and Computer-Assisted Intervention}.\hskip 1em plus 0.5em minus 0.4em\relax Springer, 2019, pp. 48--56.

\bibitem{qian2024drac}
B.~Qian \emph{et~al.}, ``{DRAC} 2022: A public benchmark for diabetic retinopathy analysis on ultra-wide optical coherence tomography angiography images,'' \emph{Patterns}, 2024.

\bibitem{odir2019peking}
G.~Challenge, ``Peking university international competition on ocular disease intelligent recognition (odir-2019),'' 2019.

\bibitem{patrini2017making}
G.~Patrini, A.~Rozza, A.~Krishna~Menon, R.~Nock, and L.~Qu, ``Making deep neural networks robust to label noise: A loss correction approach,'' in \emph{Proceedings of the IEEE Conference on Computer Vision and Pattern Recognition}, 2017, pp. 1944--1952.

\bibitem{hendrycks2018using}
D.~Hendrycks, M.~Mazeika, D.~Wilson, and K.~Gimpel, ``Using trusted data to train deep networks on labels corrupted by severe noise,'' \emph{Advances in Neural Information Processing Systems}, vol.~31, 2018.

\bibitem{cheng2022instance}
D.~Cheng \emph{et~al.}, ``Instance-dependent label-noise learning with manifold-regularized transition matrix estimation,'' in \emph{Proceedings of the IEEE/CVF Conference on Computer Vision and Pattern Recognition}, 2022, pp. 16\,630--16\,639.

\bibitem{cheng2022class}
D.~Cheng \emph{et~al.}, ``Class-dependent label-noise learning with cycle-consistency regularization,'' \emph{Advances in Neural Information Processing Systems}, vol.~35, pp. 11\,104--11\,116, 2022.

\bibitem{tanaka2018joint}
D.~Tanaka, D.~Ikami, T.~Yamasaki, and K.~Aizawa, ``Joint optimization framework for learning with noisy labels,'' in \emph{Proceedings of the IEEE Conference on Computer Vision and Pattern Recognition}, 2018, pp. 5552--5560.

\bibitem{yi2019probabilistic}
K.~Yi and J.~Wu, ``Probabilistic end-to-end noise correction for learning with noisy labels,'' in \emph{Proceedings of the IEEE/CVF Conference on Computer Vision and Pattern Recognition}, 2019, pp. 7017--7025.

\bibitem{malach2017decoupling}
E.~Malach and S.~Shalev-Shwartz, ``Decoupling ``when to update" from ``how to update",'' \emph{Advances in Neural Information Processing Systems}, vol.~30, 2017.

\bibitem{nishi2021augmentation}
K.~Nishi, Y.~Ding, A.~Rich, and T.~Hollerer, ``Augmentation strategies for learning with noisy labels,'' in \emph{Proceedings of the IEEE/CVF Conference on Computer Vision and Pattern Recognition}, 2021, pp. 8022--8031.

\bibitem{ortego2021multi}
D.~Ortego, E.~Arazo, P.~Albert, N.~E. O'Connor, and K.~McGuinness, ``Multi-objective interpolation training for robustness to label noise,'' in \emph{Proceedings of the IEEE/CVF Conference on Computer Vision and Pattern Recognition}, 2021, pp. 6606--6615.

\bibitem{liu2024plremix}
X.~Liu, B.~Zhou, and C.~Cheng, ``{PLReMix}: Combating noisy labels with pseudo-label relaxed contrastive representation learning,'' \emph{arXiv preprint arXiv:2402.17589}, 2024.

\bibitem{sun2020crssc}
Z.~Sun, X.-S. Hua, Y.~Yao, X.-S. Wei, G.~Hu, and J.~Zhang, ``{CRSSC}: Salvage reusable samples from noisy data for robust learning,'' in \emph{Proceedings of the 28th ACM International Conference on Multimedia}, 2020, pp. 92--101.

\bibitem{shin2020strategies}
W.~Shin, J.-W. Ha, S.~Li, Y.~Cho, H.~Song, and S.~Kwon, ``Which strategies matter for noisy label classification? insight into loss and uncertainty,'' \emph{arXiv preprint arXiv:2008.06218}, 2020.

\bibitem{dgani2018training}
Y.~Dgani, H.~Greenspan, and J.~Goldberger, ``Training a neural network based on unreliable human annotation of medical images,'' in \emph{2018 IEEE 15th International Symposium on Biomedical Imaging (ISBI 2018)}.\hskip 1em plus 0.5em minus 0.4em\relax IEEE, 2018, pp. 39--42.

\bibitem{le2019pancreatic}
H.~Le, D.~Samaras, T.~Kurc, R.~Gupta, K.~Shroyer, and J.~Saltz, ``Pancreatic cancer detection in whole slide images using noisy label annotations,'' in \emph{International Conference on Medical Image Computing and Computer-Assisted Intervention}, 2019, pp. 541--549.

\bibitem{xue2019robust}
C.~Xue, Q.~Dou, X.~Shi, H.~Chen, and P.-A. Heng, ``Robust learning at noisy labeled medical images: Applied to skin lesion classification,'' in \emph{2019 IEEE 16th International Symposium on Biomedical Imaging (ISBI 2019)}.\hskip 1em plus 0.5em minus 0.4em\relax IEEE, 2019, pp. 1280--1283.

\bibitem{li2023learning}
J.~Li \emph{et~al.}, ``Learning robust classifier for imbalanced medical image dataset with noisy labels by minimizing invariant risk,'' in \emph{International Conference on Medical Image Computing and Computer-Assisted Intervention}.\hskip 1em plus 0.5em minus 0.4em\relax Springer, 2023, pp. 306--316.

\bibitem{xing2023gradient}
X.~Xing, Z.~Chen, Z.~Gao, and Y.~Yuan, ``Gradient and feature conformity-steered medical image classification with noisy labels,'' in \emph{International Conference on Medical Image Computing and Computer-Assisted Intervention}.\hskip 1em plus 0.5em minus 0.4em\relax Springer, 2023, pp. 75--84.

\bibitem{khanal2023improving}
B.~Khanal, B.~Bhattarai, B.~Khanal, and C.~A. Linte, ``Improving medical image classification in noisy labels using only self-supervised pretraining,'' in \emph{MICCAI Workshop on Data Engineering in Medical Imaging}.\hskip 1em plus 0.5em minus 0.4em\relax Springer, 2023, pp. 78--90.

\bibitem{ju2022improving}
L.~Ju \emph{et~al.}, ``Improving medical images classification with label noise using dual-uncertainty estimation,'' \emph{IEEE Transactions on Medical Imaging}, vol.~41, no.~6, pp. 1533--1546, 2022.

\bibitem{zhou2023combating}
Y.~Zhou, L.~Huang, T.~Zhou, and H.~Sun, ``Combating medical noisy labels by disentangled distribution learning and consistency regularization,'' \emph{Future Generation Computer Systems}, vol. 141, pp. 567--576, 2023.

\bibitem{goldberger2016training}
J.~Goldberger and E.~Ben-Reuven, ``Training deep neural-networks using a noise adaptation layer,'' in \emph{International Conference on Learning Representations}, 2016.

\bibitem{ren2018learning}
M.~Ren, W.~Zeng, B.~Yang, and R.~Urtasun, ``Learning to reweight examples for robust deep learning,'' in \emph{International Conference on Machine Learning}.\hskip 1em plus 0.5em minus 0.4em\relax PMLR, 2018, pp. 4334--4343.

\bibitem{arpit2017closer}
D.~Arpit \emph{et~al.}, ``A closer look at memorization in deep networks,'' in \emph{International Conference on Machine Learning}.\hskip 1em plus 0.5em minus 0.4em\relax PMLR, 2017, pp. 233--242.

\bibitem{arazo2019unsupervised}
E.~Arazo, D.~Ortego, P.~Albert, N.~O’Connor, and K.~McGuinness, ``Unsupervised label noise modeling and loss correction,'' in \emph{International Conference on Machine Learning}.\hskip 1em plus 0.5em minus 0.4em\relax PMLR, 2019, pp. 312--321.

\bibitem{yu2019does}
X.~Yu, B.~Han, J.~Yao, G.~Niu, I.~Tsang, and M.~Sugiyama, ``How does disagreement help generalization against label corruption?'' in \emph{International Conference on Machine Learning}, 2019, pp. 7164--7173.

\bibitem{richardson1997bayesian}
S.~Richardson and P.~J. Green, ``On bayesian analysis of mixtures with an unknown number of components (with discussion),'' \emph{Journal of the Royal Statistical Society Series B: Statistical Methodology}, vol.~59, no.~4, pp. 731--792, 1997.

\bibitem{dempster1977maximum}
A.~P. Dempster, N.~M. Laird, and D.~B. Rubin, ``Maximum likelihood from incomplete data via the em algorithm,'' \emph{Journal of the royal statistical society: series B (methodological)}, vol.~39, no.~1, pp. 1--22, 1977.

\bibitem{berthelot2019mixmatch}
D.~Berthelot, N.~Carlini, I.~Goodfellow, N.~Papernot, A.~Oliver, and C.~A. Raffel, ``Mixmatch: A holistic approach to semi-supervised learning,'' \emph{Advances in Neural Information Processing Systems}, vol.~32, 2019.

\bibitem{chen2021beyond}
P.~Chen, J.~Ye, G.~Chen, J.~Zhao, and P.-A. Heng, ``Beyond class-conditional assumption: A primary attempt to combat instance-dependent label noise,'' in \emph{Proceedings of the AAAI Conference on Artificial Intelligence}, vol.~35, no.~13, 2021, pp. 11\,442--11\,450.

\bibitem{wilkinson2003proposed}
C.~Wilkinson \emph{et~al.}, ``Proposed international clinical diabetic retinopathy and diabetic macular edema disease severity scales,'' \emph{Ophthalmology}, vol. 110, no.~9, pp. 1677--1682, 2003.

\bibitem{deep_reitna_enhance}
Z.~Shen, H.~Fu, J.~Shen, and L.~Shao, ``Modeling and enhancing low-quality retinal fundus images,'' \emph{IEEE Transactions on Medical Imaging}, vol.~40, no.~3, pp. 996--1006, 2020.

\bibitem{he2016resnet}
K.~He, X.~Zhang, S.~Ren, and J.~Sun, ``Deep residual learning for image recognition,'' in \emph{2016 IEEE Conference on Computer Vision and Pattern Recognition}, 2016, pp. 770--778.

\bibitem{deng2009imagenet}
J.~Deng, W.~Dong, R.~Socher, L.-J. Li, K.~Li, and L.~Fei-Fei, ``Imagenet: A large-scale hierarchical image database,'' in \emph{2009 IEEE Conference on Computer Vision and Pattern Recognition}.\hskip 1em plus 0.5em minus 0.4em\relax Ieee, 2009, pp. 248--255.

\bibitem{zhang2018mixup}
H.~Zhang, M.~Cisse, Y.~N. Dauphin, and D.~Lopez-Paz, ``mixup: Beyond empirical risk minimization,'' in \emph{International Conference on Learning Representations}, 2018.

\bibitem{kaggle}
Kaggle, ``Kaggle diabetic retinopathy detection competition.'' \url{https://www.kaggle.com/c/diabetic-retinopathy-detection}.

\bibitem{mcinnes2018umap}
L.~McInnes, J.~Healy, N.~Saul, and L.~Gro{\ss}berger, ``{UMAP}: Uniform manifold approximation and projection,'' \emph{Journal of Open Source Software}, vol.~3, no.~29, 2018.

\end{thebibliography}

\end{document}